\documentclass[letterpaper]{article}
\usepackage[preprint]{aaai2027}

\usepackage[hyphens]{url}
\usepackage{graphicx}
\definecolor{GitHubLinkBlue}{HTML}{0969DA}
\newcommand{\CodeRepositoryURL}{%
  \leavevmode
  \pdfstartlink attr{/Border [0 0 0]} user{%
    /Subtype /Link
    /A << /S /URI /URI (https://github.com/fanrj3/GeoMoE) >>%
  }%
  \mbox{%
    \raisebox{-0.12em}{\includegraphics[height=0.9em]{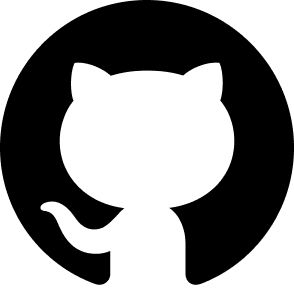}}%
    \hspace{0.2em}\textcolor{GitHubLinkBlue}{\url{github.com/fanrj3/GeoMoE}}%
  }%
  \pdfendlink
}
\usepackage{natbib}
\usepackage{caption}
\usepackage{amsmath}
\usepackage{amssymb}
\usepackage{booktabs}
\usepackage{multirow}
\usepackage{array}
\usepackage{multicol}

\title{One Query, Many Scales: Sparse Mixture-of-Experts for Efficient
Hierarchical Cross-View Geo-Localization}
\author{
  Ruijie~Fan\textsuperscript{\rm 1,\rm 2},
  Junyan~Ye\textsuperscript{\rm 2},
  Qi~Zhu\textsuperscript{\rm 2},
  Weijia~Li\textsuperscript{\rm 1}\corresponding
}
\affiliations{
  \textsuperscript{\rm 1}Tsinghua Shenzhen International Graduate School,
  Tsinghua University\\
  \textsuperscript{\rm 2}School of Geospatial Engineering and Science,
  Sun Yat-sen University\\
  liweijia@sz.tsinghua.edu.cn
}

\begin{document}

% Save the bibliography command after aaai2027 applies its begin-document hooks.
\let\ArxivBibliography\bibliography

\maketitle

\begin{abstract}
Cross-view geo-localization (CVGL) retrieves geo-tagged satellite imagery for a
ground-view query. Most systems exhaustively search a flat, fixed-resolution
gallery, incurring high cost over large areas and adapting poorly to satellite
resolution changes. Autoregressive coarse-to-fine alternatives reduce
comparisons but bind later predictions to earlier decisions and a predefined
hierarchy. We introduce GeoMoE, a sparse mixture-of-experts dual encoder that
decouples global multi-scale representation learning from local hierarchical
search. Global multi-scale supervision and content-adaptive routing map ground
and satellite images across resolutions into a globally comparable embedding
space. At inference, each image is encoded once, and probabilistic beam search
follows parent--child links to score a small candidate subset. Later levels
reuse these descriptors rather than features generated by preceding levels,
limiting feature-level error propagation and hierarchy coupling. We further
introduce VIGOR-M, a four-city benchmark with an explicit parent--child
satellite hierarchy and held-out half-step galleries for single-resolution,
cross-resolution, and hierarchical evaluation. GeoMoE achieves 95.78\% R@40m
on Just Zoom In, 2.77 percentage points above the previous best, and 62.39\%
R@1 on VIGOR-M. The latter requires 0.885 MMAC/query for descriptor matching,
5.27\% of an exhaustive L3 scan, while exceeding the strongest exhaustive
baseline by 3.12 percentage points in R@1. One model trained on L1,
L2, and L3 also outperforms a matched dense control across all six galleries
and transfers to three withheld resolutions. By decoupling globally trained
embeddings from local hierarchical search, GeoMoE jointly improves localization
accuracy, search efficiency, and cross-resolution transfer. Code is available at
\CodeRepositoryURL.
\end{abstract}

\section{Introduction}

Cross-view geo-localization (CVGL) estimates the geographic location of a ground-view image by retrieving matching imagery from a geo-tagged satellite gallery \cite{WIGL}. It supports localization when geospatial metadata or reliable global navigation satellite system (GNSS) signals are unavailable and connects ground-level observations with large-scale overhead context. With expanding high-resolution satellite imagery and street-view coverage, the task has become increasingly practical. Yet CVGL remains challenging because ground and satellite images observe the same place from markedly different viewpoints. Ground images emphasize facades, roads, and local semantics, whereas satellite images reveal rooftops and spatial layouts. A practical system must therefore bridge this discrepancy, generalize across regions, and search large reference galleries efficiently.

\begin{figure}[t]
    \centering
    \includegraphics[width=\columnwidth]{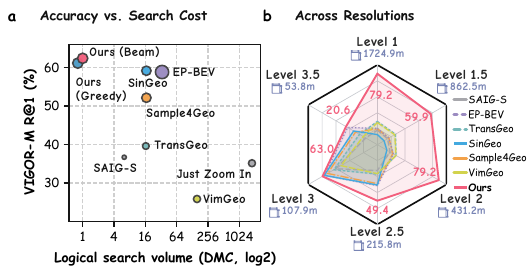}
    \caption{GeoMoE on VIGOR-M. (a) R@1 versus descriptor-matching computations
    (DMC, MMAC/query; log$_2$ scale), where upper left is better. (b) Single-resolution R@1
    across six levels and their representative ground-coverage widths. GeoMoE
    is trained on L1/L2/L3; half-step levels are withheld, and baselines use
    L3-trained models.}
    \label{fig:flag}
\end{figure}

Existing approaches usually learn a shared embedding space and rank fixed-resolution satellite crops for each ground query. Shared representations and transformer-based models have substantially improved retrieval accuracy \cite{CVMNet,TransGeo}. Despite these advances, most methods treat satellite crops as independent candidates, leaving spatial adjacency and cross-resolution parent--child relations unused. Current benchmarks reinforce this flat formulation, as VIGOR extends one-to-one matching with overlapping crops but still provides discrete candidates at a single resolution \cite{VIGOR}. For an $N \times N$ tiling of a target region, the gallery grows as $O(N^2)$, and exhaustive retrieval compares each query against all candidates. Scaling geographic coverage therefore increases both storage and similarity-computation costs.

Recent work has begun to explore hierarchical formulations for large-scale and fine-grained CVGL by integrating retrieval with metric localization \cite{UnifyGeo}. More recently, Just Zoom In constructs a multi-resolution satellite hierarchy and retrieves a target cell through autoregressive coarse-to-fine zooming \cite{justzoomin}. Although this formulation reduces the number of evaluated candidates, later decisions are conditioned on earlier selections and can therefore be vulnerable to error propagation: an incorrect coarse-level decision may exclude the correct region from subsequent stages. Its behavior is also tied to the predefined zoom hierarchy and terminal resolution, limiting flexibility across hierarchy configurations and deployment scales.

We propose \textbf{GeoMoE}, a sparse mixture-of-experts dual encoder for efficient and resolution-adaptive cross-view geo-localization. Multiple lightweight experts and a learnable router model representations across spatial resolutions. A unified embedding space allows each ground or satellite image to be encoded once and compared directly across views and resolutions.

GeoMoE follows a \emph{global discrimination, local execution} principle. During training, it learns globally discriminative representations using samples from the complete reference space, thereby preserving strong retrieval performance at each individual resolution level. During inference, a lightweight learnable beam-search module exploits the spatial hierarchy of the satellite gallery and evaluates only a small subset of indexed candidates. This design combines the representation quality of global retrieval with the efficiency of coarse-to-fine search, without introducing autoregressive feature dependencies between hierarchical levels. Figure~\ref{fig:flag} summarizes its accuracy--efficiency trade-off and performance across satellite resolutions.

Our main contributions are summarized as follows:
\begin{itemize}
    \item We introduce an index-based hierarchical retrieval paradigm based on \emph{global discrimination and local execution}. A single model supports both exhaustive retrieval at any individual resolution and efficient coarse-to-fine retrieval across multiple levels, while remaining robust to changes in hierarchy and resolution configurations.

    \item We propose \textbf{GeoMoE}, a sparse and lightweight mixture-of-experts dual encoder for single-pass, multi-resolution cross-view representation learning. GeoMoE jointly encodes ground-view images and satellite images from different hierarchical levels into a unified embedding space, enabling direct retrieval across views and resolutions.

    \item We introduce \textbf{VIGOR-M}, a large-scale, multi-city, and multi-resolution benchmark for hierarchical CVGL. It extends the original benchmark with spatially continuous satellite imagery at multiple resolutions and supports both cross-resolution and hierarchical retrieval evaluation.

    \item Under an explicit resolution-shift protocol, a single GeoMoE model trained at L1, L2, and L3 transfers to three withheld resolution levels. A matched dense MLJ control separates the contribution of sparse routing from that of multi-scale supervision.
\end{itemize}

\begin{figure*}[t]
    \centering
    \includegraphics[width=\textwidth]{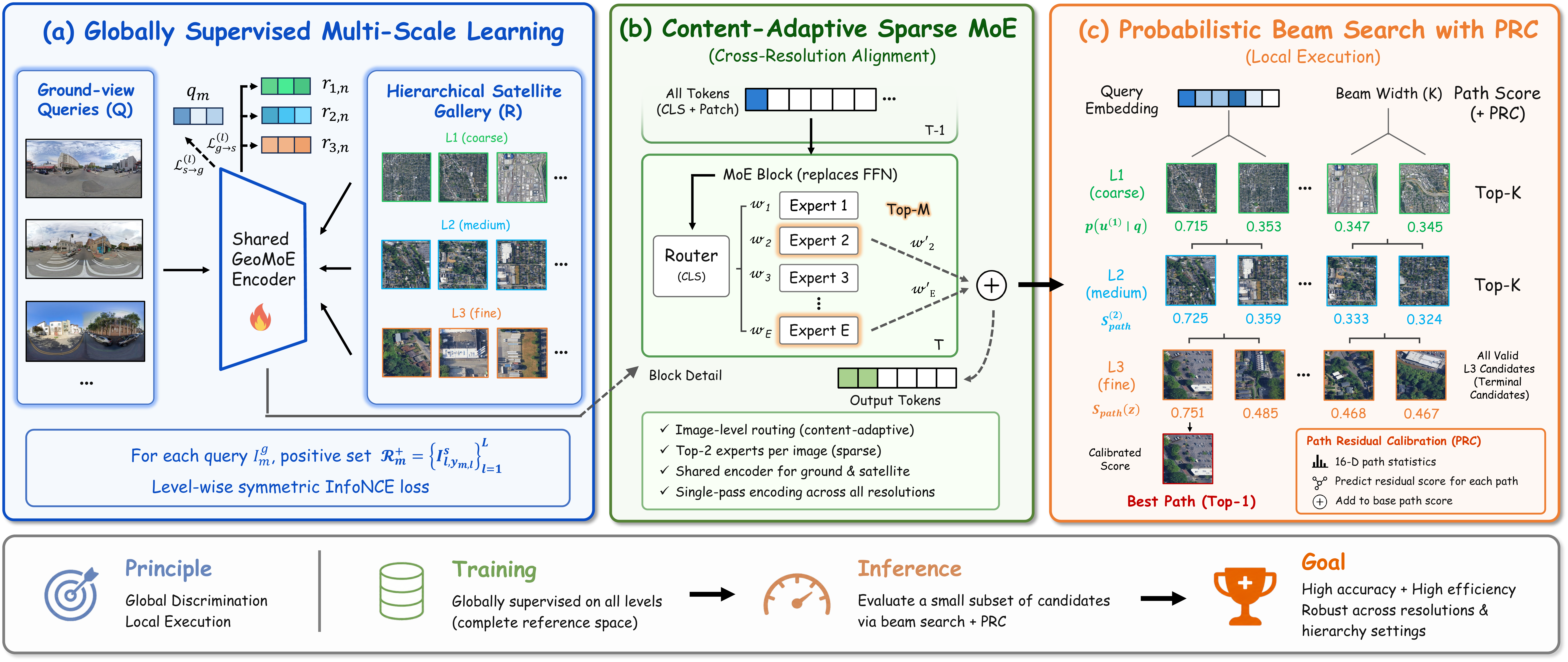}
    \caption{Overview of GeoMoE. (a) Global multi-scale supervision applies
    level-wise symmetric InfoNCE over complete galleries. (b) Image-level Top-2
    routing activates sparse experts under retrieval and load-balancing
    objectives. (c) Probabilistic beam search traverses parent--child links over
    precomputed embeddings, and PRC reranks terminal candidates.}
    \label{fig:pipeline}
\end{figure*}

\section{Related Work}

\subsection{Cross-View Geo-Localization}

Cross-view geo-localization takes a ground-view image as a query and retrieves
its location from a geo-tagged satellite gallery. Early work learned cross-view
feature translations \cite{Lin2013,WIGL}, followed by Siamese and triplet
retrieval objectives \cite{VoHays2016,CVMNet} and building-level urban matching
\cite{Tian2017}. Beyond gallery retrieval, SNAP learns self-supervised neural
maps from ground-level and overhead imagery \cite{SNAP}.

Subsequent methods incorporate orientation and alignment
\cite{LendingOrientation,DSM2020,Revisiting2021}, spatial aggregation and
feature transport \cite{SAFA,CVFT}, and bird's-eye or orthogonal-view geometry
\cite{PanoramaBEV,OVA2024}. Cross-view synthesis
\cite{Regmi2018,Toker2021,SkyDiffusion2025,FineGrainedStreet2Sat2025,
CrossViewDiff2024,Geo2}, geometric layout modeling
\cite{GeoDTR,GeoDTRPlus,FRGeo}, and semantic anchors \cite{GeoBridge} further
reduce view discrepancies, while temporal augmentation and unlabeled training
address scene changes and annotation dependence \cite{Rodrigues2021,UCVGL2024}.
Attention-based aggregation captures global layout \cite{L2LTR,TransGeo},
whereas hard-negative mining and cross-view consistency improve discrimination
under field-of-view variations
\cite{Sample4Geo,ConGeo,SinGeo}, but generally retain fixed-resolution galleries.
UnifyGeo combines regional retrieval with metric localization \cite{UnifyGeo},
whereas Just Zoom In traverses a multi-resolution hierarchy \cite{justzoomin}.
GeoMoE instead unifies resolution-specialized representations for both
single-resolution and coarse-to-fine retrieval.

\subsection{Mixture-of-Experts Models}

Mixture-of-experts (MoE) models route inputs among expert subnetworks
\cite{MoEExperts}. Sparse gating limits activation \cite{SparselyGatedMoE},
while constrained or expert-selected routing improves load balance and stability
\cite{SwitchTransformer,ExpertChoice}. Automatic sharding, balanced assignment,
and hash routing address scaling bottlenecks \cite{GShard,BASELayers,HashLayers}.
Vision and multimodal variants route patches or modalities through sparse or
soft experts \cite{VMoE,SoftMoE,LIMoE}. In geo-localization, experts have been
specialized by ground-view field of view \cite{HCLGeo}, used for key--value
feature aggregation \cite{KVRouteGeo}, or routed over UAV--satellite grids
\cite{SMGeo}. GeoMoE instead specializes experts by satellite resolution for
hierarchical ground-to-satellite retrieval.

\subsection{Cross-View Geo-Localization Datasets}

CVUSA and CVACT pair panoramas with fixed-resolution satellite images
\cite{WIGL,LendingOrientation}, while VIGOR permits multiple overlapping
candidates but remains single-resolution \cite{VIGOR}. University-1652 and
SUES-200 introduce satellite--UAV--ground matching and altitude variation
\cite{University1652,SUES200}; CVGlobal, UniGeoRS, and CVG-Text broaden city,
view, and query modalities \cite{PanoramaBEV,UniGeoRS,WhereAmI2025}. None
organizes the same area into a multi-resolution parent--child gallery. VIGOR-M
provides this hierarchy for single-resolution, cross-resolution, and
coarse-to-fine retrieval.

\section{Methodology}

GeoMoE jointly addresses cross-resolution representation learning and
efficient hierarchical retrieval in cross-view geo-localization
(Figure~\ref{fig:pipeline}). Globally supervised multi-level joint training
(MLJ) decouples representation learning from a predefined retrieval hierarchy.
A content-adaptive sparse mixture-of-experts network aligns ground-view and
satellite representations across spatial resolutions, while probabilistic beam
search identifies fine-level tiles from a subset of the reference gallery.

\subsection{Global Multi-Scale Hierarchical Retrieval}
\label{sec:global_retrieval}

\paragraph{Global Multi-Level Supervision.}

Let \(\mathcal{Q}=\{I_m^g\}_{m=1}^{M}\) be \(M\) ground-view RGB queries of
size \(H_g\times W_g\), and let
\(\mathcal{R}=\{\mathcal{R}_l\}_{l=1}^{L}\) be a hierarchical satellite
gallery. Level \(\mathcal{R}_l\) contains \(N_l\) RGB tiles \(I_{l,n}^s\) of
size \(H_l\times W_l\). If \(y_{m,l}\) indexes the positive tile for query
\(I_m^g\) at level \(l\), then
\(\mathcal{R}_m^+=\{I_{l,y_{m,l}}^s\}_{l=1}^{L}\).

For \(l\geq2\), let \(\mathcal{C}_l(u)\subseteq\mathcal{R}_l\) denote the valid
next-level nodes reachable from \(u\). In a strictly nested gallery, this
relation is induced by a parent map
\(\pi_l:\mathcal{R}_l\rightarrow\mathcal{R}_{l-1}\); the set-valued definition
also accommodates overlapping proposal neighborhoods at the root level. Rather
than constructing supervision only within descendants selected at preceding
levels, positive pairs and negatives span the complete geographic extent of
each level. Thus,
\(\mathcal{D}_l=\{(I_m^g,I_{l,y_{m,l}}^s,l)\}_{m=1}^{M}\), and the complete
multi-scale training set is \(\mathcal{D}=\bigcup_{l=1}^{L}\mathcal{D}_l\). This
global formulation decouples representation learning from a particular
hierarchy design and supports different hierarchy depths and gallery sizes
without modifying the objective or retraining the model.

\paragraph{Level-Wise Symmetric InfoNCE.}

Satellite tiles from different hierarchy levels may cover the same geographic
region while exhibiting substantially different spatial resolutions and
contextual ranges. Directly mixing samples from all levels within a single
contrastive denominator may therefore introduce semantically conflicting
negatives. To avoid such conflicts, we adopt a level-wise symmetric InfoNCE
objective together with balanced multi-scale sampling.

Suppose that a mini-batch contains \(B_l\) positive pairs from level \(l\),
with normalized ground-view and satellite representations
\(\{\mathbf{q}_i^{(l)}\}_{i=1}^{B_l}\) and
\(\{\mathbf{r}_i^{(l)}\}_{i=1}^{B_l}\), respectively. We define the pairwise
similarity logit as
\(A_{ij}^{(l)}=(\mathbf{q}_i^{(l)})^{\top}\mathbf{r}_j^{(l)}/\tau\), where
\(\tau\) is a temperature hyperparameter.

The ground-to-satellite and satellite-to-ground contrastive losses are
\begin{equation}
\begin{aligned}
\mathcal{L}_{g\rightarrow s}^{(l)}
&=
-\frac{1}{B_l}
\sum_{i=1}^{B_l}
\log
\frac{
\exp\left(A_{ii}^{(l)}\right)
}{
\sum_{j=1}^{B_l}
\exp\left(A_{ij}^{(l)}\right)
}, \\
\mathcal{L}_{s\rightarrow g}^{(l)}
&=
-\frac{1}{B_l}
\sum_{i=1}^{B_l}
\log
\frac{
\exp\left(A_{ii}^{(l)}\right)
}{
\sum_{j=1}^{B_l}
\exp\left(A_{ji}^{(l)}\right)
}.
\end{aligned}
\end{equation}

The level-wise loss is
\(\mathcal{L}_{\mathrm{ret}}^{(l)}=(\mathcal{L}_{g\rightarrow s}^{(l)}+
\mathcal{L}_{s\rightarrow g}^{(l)})/2\). We aggregate levels as
\(\mathcal{L}_{\mathrm{ret}}=(\sum_l w_l)^{-1}\sum_l
w_l\mathcal{L}_{\mathrm{ret}}^{(l)}\), where \(w_l\) is the weight of level
\(l\).

\subsection{Content-Adaptive Sparse MoE Network}
\label{sec:geome_architecture}

\paragraph{GeoMoE Block and Image-Level Routing.}

Satellite resolution changes both the available visual evidence and its
spatial context: coarse images emphasize urban layouts, road networks, and
regional semantics, whereas fine images emphasize individual buildings and
local structures. Independent models incur substantial computation and storage
and limit cross-scale knowledge sharing, whereas dense sharing can cause
interference among scale-specific objectives. GeoMoE therefore shares a common
representation space while sparsely activating specialized feature
transformations for each image.

We follow a pre-normalization Transformer architecture and replace the dense
feed-forward network (FFN) with a sparse MoE layer, resulting in the proposed
GeoMoE block. Let
\(\mathbf{X}\in\mathbb{R}^{(T+1)\times d}\) denote an input sequence containing
one class token (\(\mathrm{CLS}\)) and \(T\) patch tokens, where \(d\) is the feature
dimension. Using multi-head self-attention (MSA) and layer normalization (LN),
a GeoMoE block is formulated as
\begin{equation}
\begin{aligned}
\widetilde{\mathbf{X}}
&=
\mathbf{X}
+
\operatorname{MSA}
\left(
\operatorname{LN}_1(\mathbf{X})
\right), \\
\mathbf{X}'
&=
\widetilde{\mathbf{X}}
+
\operatorname{MoE}
\left(
\operatorname{LN}_2(\widetilde{\mathbf{X}})
\right).
\end{aligned}
\end{equation}

For the \(i\)-th image in a mini-batch, we extract its normalized
\(\mathrm{CLS}\) representation as the routing feature
\(\mathbf{h}_i=[\operatorname{LN}_2(\widetilde{\mathbf{X}}_i)]_{\mathrm{CLS}}
\in\mathbb{R}^{d}\). A lightweight linear router predicts logits
\(\mathbf{a}_i=\mathbf{W}_r\mathbf{h}_i+\mathbf{b}_r\) over \(E\) experts,
followed by \(\mathbf{p}_i=\operatorname{softmax}(\mathbf{a}_i)\), where
\(\mathbf{p}_i\in\mathbb{R}^{E}\).

Following standard sparse MoE designs, we employ Top-\(k\) routing. Let
\(\mathcal{T}_k(\mathbf{p}_i)\) denote the indices of the \(k\) experts with the
highest routing probabilities for image \(i\). The routing weights of the
selected experts are renormalized as
\(\alpha_{i,e}=p_{i,e}/\sum_{j\in\mathcal{T}_k(\mathbf{p}_i)}p_{i,j}\) for
\(e\in\mathcal{T}_k(\mathbf{p}_i)\). We use \(k=2\), so only two experts are
activated for each image. Given \(\mathbf{Z}_i=\operatorname{LN}_2(\widetilde{\mathbf{X}}_i)\), the MoE output is
\begin{equation}
\operatorname{MoE}(\mathbf{Z}_i)
=
\sum_{e\in\mathcal{T}_k(\mathbf{p}_i)}
\alpha_{i,e}
\operatorname{FFN}_e(\mathbf{Z}_i),
\end{equation}
where \(\operatorname{FFN}_e\) denotes the \(e\)-th expert network. Image-level
routing uses the global \(\mathrm{CLS}\) representation to select experts that
transform the token sequence. Expert selection therefore reflects
global image content and spatial-resolution characteristics without introducing
token-level routing instability.

\paragraph{Load-Balancing Loss and Training Objective.}

Without explicit regularization, the router may concentrate most images on a
small subset of experts, resulting in expert collapse. GeoMoE therefore adopts
a Switch-style objective to encourage balanced expert utilization.

For a mini-batch of \(B\) images, the average soft routing probability assigned
to expert \(e\) is \(\overline{p}_e=B^{-1}\sum_{i=1}^{B}p_{i,e}\). The
normalized hard routing load after top-\(k\) selection is
\(\overline{m}_e=(kB)^{-1}\sum_{i=1}^{B}\mathbb{I}
[e\in\mathcal{T}_k(\mathbf{p}_i)]\), where \(\mathbb{I}[\cdot]\) is the
indicator function. The load-balancing loss is formulated as
\(\mathcal{L}_{\mathrm{balance}}=
E\sum_{e=1}^{E}\overline{p}_e\overline{m}_e\). When both the soft routing
probabilities and hard expert assignments are uniformly distributed among all
experts, this loss approaches \(1\); it penalizes the joint concentration of
routing probability and assignment frequency. The complete training objective
combines the multi-scale retrieval and auxiliary load-balancing losses,
\(\mathcal{L}=\mathcal{L}_{\mathrm{ret}}+
\lambda_{\mathrm{aux}}\mathcal{L}_{\mathrm{balance}}\), where
\(\lambda_{\mathrm{aux}}\) controls the contribution of the load-balancing
regularization.

% Register the main comparison before the final method subsection so it can
% occupy the next double-column page top.
\begin{table*}[!t]
\centering
{\small
\renewcommand{\arraystretch}{1.10}
\setlength{\tabcolsep}{1.5pt}
\begin{tabular*}{\textwidth}{@{\extracolsep{\fill}}lcccccccccc@{}}
\toprule
\multirow{2}{*}{Method}
& \multicolumn{5}{c}{Just Zoom In}
& \multicolumn{5}{c}{VIGOR-M} \\
\cmidrule(lr){2-6}
\cmidrule(lr){7-11}
& R@1 $\uparrow$
& R@40m $\uparrow$
& R@50m $\uparrow$
& R@100m $\uparrow$
& Median (m) $\downarrow$
& R@1 $\uparrow$
& R@100m $\uparrow$
& R@200m $\uparrow$
& R@300m $\uparrow$
& Median (m) $\downarrow$ \\
\midrule

TransGeo {\scriptsize (CVPR'22)}
& 61.01 & 82.17 & 85.67 & 92.17 & 19.37
& 39.60 & 50.07 & 57.85 & 62.03 & 99.55 \\

SAIG-D {\scriptsize (arXiv'23)}
& 61.34 & 78.83 & 81.33 & 86.84 & 19.18
& 36.69 & 46.69 & 53.92 & 57.70 & 133.36 \\

Sample4Geo {\scriptsize (ICCV'23)}
& 64.77 & 82.32 & 84.98 & 89.80 & 18.64
& 52.14 & 63.56 & 70.50 & 74.06 & 60.65 \\

EP-BEV {\scriptsize (ECCV'24)}
& 65.67 & 83.83 & 86.48 & 91.71 & 18.47
& 58.82 & 71.14 & 77.62 & \underline{80.70} & 54.20 \\

VimGeo {\scriptsize (IJCAI'25)}
& 44.74 & 63.87 & 68.71 & 80.76 & 25.34
& 27.74 & 35.88 & 44.33 & 49.92 & 301.67 \\

SinGeo {\scriptsize (CVPR'26)}
& \underline{79.95}
& \underline{93.01}
& \underline{93.63}
& \underline{95.00}
& \underline{16.61}
& \underline{59.27}
& \underline{72.54}
& \underline{77.94}
& 80.21 & \underline{53.95} \\

Just Zoom In {\scriptsize (arXiv'26)}
& 70.55
& 90.16
& 91.57
& 94.66
& 17.57
& 35.13 & 64.08 & 74.79 & 79.37 & 65.61 \\

\textbf{GeoMoE (Ours)}
& \textbf{81.71}
& \textbf{95.78}
& \textbf{96.32}
& \textbf{97.68}
& \textbf{16.43}
& \textbf{62.39}
& \textbf{77.82}
& \textbf{84.30}
& \textbf{86.86}
& \textbf{52.33} \\

\bottomrule
\end{tabular*}
}
\caption{Hierarchical localization performance on Just Zoom In and VIGOR-M.
Recall values are percentages, and Median is the median top-1 localization
error in meters. Bold and underlined values indicate the best and second-best
results, respectively.}
\label{tab:vigor_m_results}
\end{table*}

\subsection{Probabilistic Coarse-to-Fine Beam Search}
\label{sec:beam_search}

\paragraph{Hierarchical Path Scoring.}

Exhaustive retrieval over the finest-resolution gallery requires comparing a
query against every fine-level satellite descriptor. GeoMoE instead exploits
parent--child relations, scoring only descendants of retained hypotheses while
maintaining multiple beam paths. This reduces fine-level comparisons and limits
cascading errors from premature parent selection.

Given a retained parent node \(u\) at level \(l-1\), we normalize the
similarity scores only over its valid child set \(\mathcal{C}_l(u)\). The
conditional probability of selecting a child \(v\) is defined as
\begin{equation}
P(v\mid u,\mathbf{q})
=
\frac{
\exp\left(s_l(v)/\tau\right)
}{
\sum_{k\in\mathcal{C}_l(u)}
\exp\left(s_l(k)/\tau\right)
},
\qquad
v\in\mathcal{C}_l(u),
\end{equation}
where \(s_l(v)\) denotes the similarity between query representation
\(\mathbf{q}\) and satellite node \(v\) at level \(l\). For an arbitrary hierarchical path \(z=(v_1,v_2,\ldots,v_j)\), its conditional
probability can be factorized as
\begin{equation}
P(z\mid\mathbf{q})
=
P(v_1\mid\mathbf{q})
\prod_{l=2}^{j}
P(v_l\mid v_{l-1},\mathbf{q}).
\end{equation}

For numerical stability, we compute the corresponding log-probability:
\begin{equation}
\begin{aligned}
S_{\mathrm{path}}(z)
&=
\log P(z\mid\mathbf{q}) \\
&=
\log P(v_1\mid\mathbf{q})
+
\sum_{l=2}^{j}
\log
P(v_l\mid v_{l-1},\mathbf{q}).
\end{aligned}
\end{equation}
At each nonterminal level, all descendants of the retained paths are scored.
Nodes reached through multiple paths are merged by retaining the maximum
accumulated path score before the globally highest-scoring \(K_p\) distinct
paths are preserved. At the terminal level, all valid children of the retained
penultimate nodes form the complete candidate set passed to PRC for final
ranking.

\paragraph{Path Residual Calibration (PRC).}

Local normalization makes child scores comparable within a parent, but not
necessarily across parents with different branching factors, score
distributions, or conditional entropies. Directly accumulating conditional
log-probabilities can therefore introduce systematic biases among paths from
different local subtrees.

To compensate for these biases, a lightweight residual calibrator reranks the
beam candidates. For each path \(z\), a 16-dimensional descriptor
\(\widehat{\boldsymbol{\phi}}(z)\in\mathbb{R}^{16}\) summarizes its multi-level
confidence and structural statistics. A lightweight multilayer
perceptron predicts a residual correction
\(g_{\psi}(\widehat{\boldsymbol{\phi}}(z))\), yielding the calibrated path score
\(S_{\mathrm{cal}}(z)=S_{\mathrm{path}}(z)+
g_{\psi}(\widehat{\boldsymbol{\phi}}(z))\). The residual formulation preserves
the beam-search score as the primary ranking signal while correcting systematic
path-dependent biases. The calibrator is fitted on the official training split,
with an internal training partition reserved for checkpoint selection; official
test labels are not used for either step. Descriptor construction and calibrator
training are provided in the supplementary material.

\FloatBarrier
\section{The VIGOR-M Benchmark}
\label{sec:vigor_m}

\begin{figure}[!t]
    \centering
    \includegraphics[width=\columnwidth]{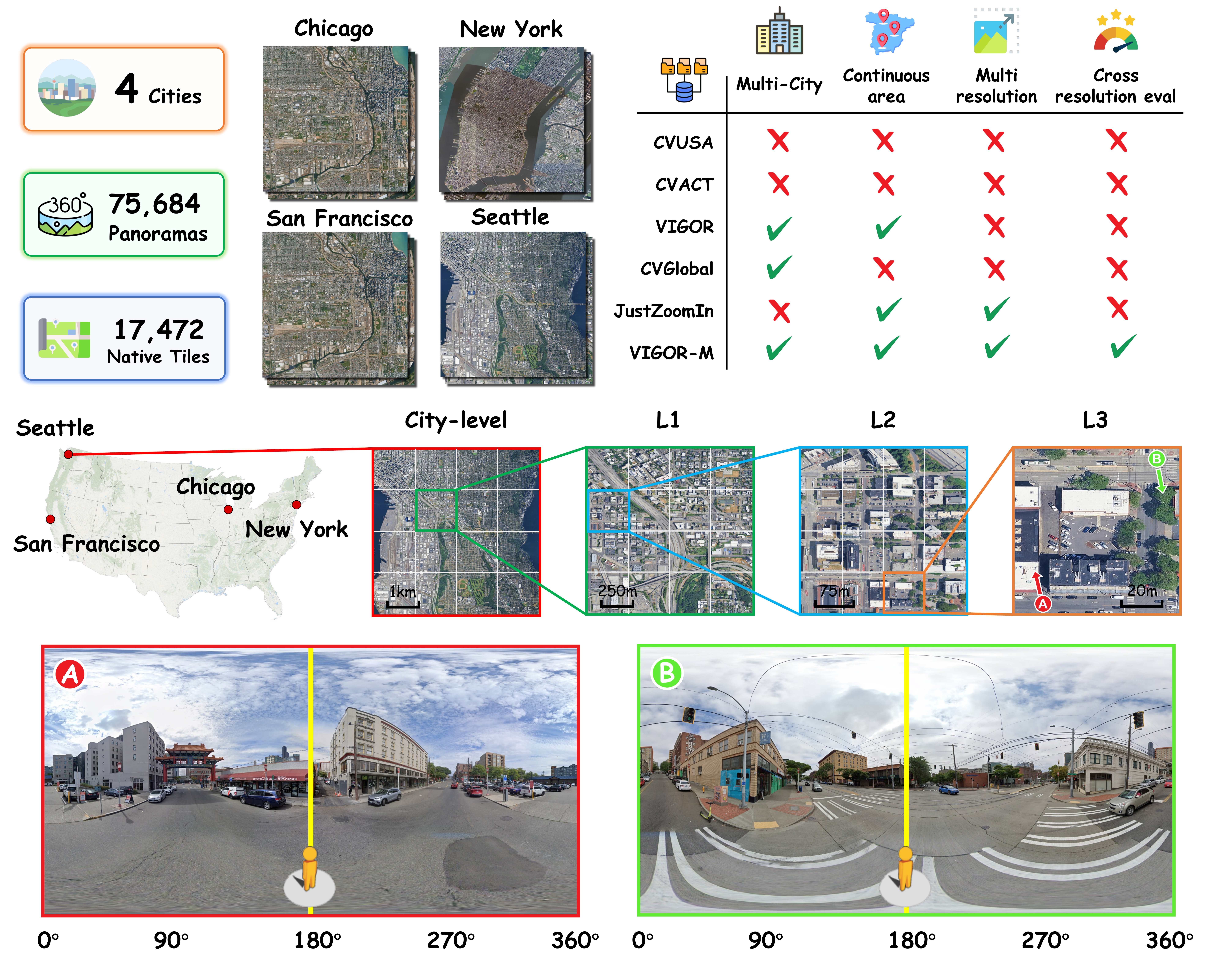}
    \caption{Geographic coverage and multi-resolution hierarchy of VIGOR-M.
    Four city maps are recursively partitioned into nested L1--L3 tiles. Scale
    bars indicate representative ground coverage, and panoramas A and B show
    associated ground queries.}
    \label{fig:vigor_m_dataset}
\end{figure}

We construct \textbf{VIGOR-M}, a multi-scale benchmark for hierarchical
cross-view geo-localization over continuous urban areas. It extends the
geographic coverage of VIGOR to a nested satellite hierarchy spanning
New York, San Francisco, Chicago, and Seattle, with 75,684 ground panoramas
collected from Street View Download 360. Figure~\ref{fig:vigor_m_dataset}
summarizes the geographic coverage, hierarchy, and query-to-tile
correspondence.

VIGOR-M is designed for map-specific localization over a predefined urban
gallery, reflecting deployments in which the service region and reference map
are known. Following the VIGOR same-area protocol \cite{VIGOR}, training and
test ground queries are identity-disjoint but spatially interleaved across the
same four city maps. All 64 native L1 tiles and 995 of 998 occupied L2 tiles
occur in both splits; 93.23\% of test queries map to L3 positives observed
during training. Inference nevertheless searches the complete four-city gallery
without a city oracle. The reported results therefore measure localization and
cross-resolution transfer within known maps, rather than generalization to
unseen cities. Each panorama is assigned a random north-orientation offset,
while its camera orientation is retained for orientation-aware analysis.

Each city extent is padded to a square region and recursively partitioned into
a \(4\times4\) grid at L1, L2, and L3. Across four cities, level \(l\) contains
\(N_l=64\times16^{l-1}\) tiles for \(l\in\{1,2,3\}\), yielding 64, 1,024, and
16,384 native tiles, respectively. Three half-step galleries are constructed
in parallel: L1.5 partitions each city into an \(8\times8\) grid, and recursive
\(4\times4\) subdivision produces L2.5 and L3.5. Their ground-coverage widths
are half those at L1, L2, and L3, respectively. All satellite images are resized
to \(784\times784\) pixels, while panoramas use \(2048\times1024\) pixels. For
retrieval, each overlapping L1 root \(u\) expands to a \(4\times4\) L2 proposal
set \(\mathcal{C}_2(u)\), so an L2 node may be reached through multiple roots.
Index construction and DMC accounting are detailed in the supplementary
material. The half-step galleries are excluded from training and used
exclusively to evaluate within-map transfer to unseen resolutions.

\begin{figure*}[t]
    \centering
    \includegraphics[width=\textwidth]{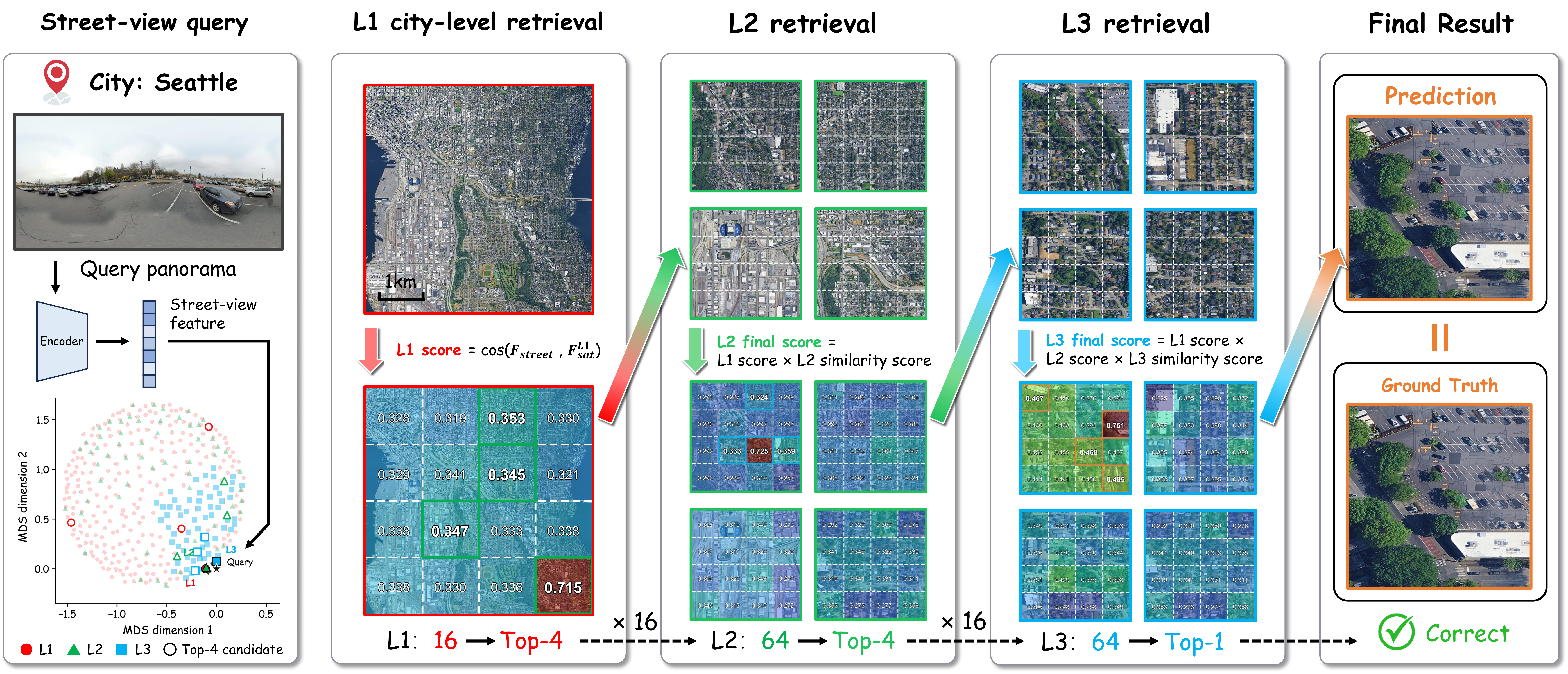}
    \caption{Qualitative hierarchical retrieval on VIGOR-M for a correctly
    localized query. The left panel shows the query panorama and a
    two-dimensional projection of level-specific embeddings obtained by
    multidimensional scaling (MDS). GeoMoE retains the top four paths at L1 and
    L2, then selects the top-ranked L3 tile. Numeric labels show cosine
    similarities at L1 and cumulative path scores at later levels. The final
    prediction matches the ground-truth tile.}
    \label{fig:qualitative_results}
\end{figure*}

\begin{table}[!t]
\centering
{\small
\renewcommand{\arraystretch}{1.12}
\setlength{\tabcolsep}{1.5pt}
\begin{tabular*}{\columnwidth}{@{\extracolsep{\fill}}lcccccc@{}}
\toprule
Method & L1 & L1.5 & L2 & L2.5 & L3 & L3.5 \\
Tile width (m) & 1724.9 & 862.5 & 431.2 & 215.8 & 107.9 & 53.8 \\
\midrule
\multicolumn{7}{l}{\textit{Standard single-scale training (L3)}} \\
TransGeo~\shortcite{TransGeo} & 8.47 & 6.91 & 7.37 & 13.61 & 39.60 & 6.71 \\
SAIG-S~\shortcite{SAIG} & 3.31 & 2.20 & 3.38 & 10.57 & 36.69 & 2.87 \\
Sample4Geo~\shortcite{Sample4Geo} & 3.96 & 2.80 & 4.99 & 23.99 & 52.14 & 10.00 \\
EP-BEV~\shortcite{PanoramaBEV} & 4.16 & 2.83 & 5.32 & 26.40 & 58.82 & \underline{18.61} \\
VimGeo~\shortcite{VimGeo} & 7.70 & 6.30 & 7.30 & 13.20 & 27.74 & 4.20 \\
SinGeo~\shortcite{SinGeo} & 1.61 & 0.95 & 1.89 & 25.56 & \underline{59.27} & 11.80 \\
\midrule
\multicolumn{7}{l}{\textit{Joint multi-scale training (L1/L2/L3)}} \\
MLJ (Dense) & \underline{48.11} & \underline{59.35} & \underline{71.43} & \underline{45.39} & 51.67 & 8.24 \\
\textbf{GeoMoE (Ours)} & \textbf{79.22} & \textbf{59.86} & \textbf{79.20} & \textbf{49.40} & \textbf{63.01} & \textbf{20.63} \\
\bottomrule
\end{tabular*}
}
\caption{Cross-resolution R@1 (\%) on VIGOR-M. Standard baselines are trained on
L3, whereas MLJ (Dense) and GeoMoE use L1/L2/L3; half-step levels are withheld.
Tile width denotes ground coverage. Bold and underlined values mark the best and
second-best results.}
\label{tab:single_resolution_results}
\end{table}

% Register the remaining experiment tables before the section body so LaTeX
% can distribute them across the next two-column pages.
\begin{table}[t]
\centering
{\small
\renewcommand{\arraystretch}{1.04}
\setlength{\tabcolsep}{1.8pt}
\begin{tabular*}{\columnwidth}{@{\extracolsep{\fill}}lrr@{\hspace{4pt}}lrr@{}}
\toprule
Method & DMC & R@1 &
Method & DMC & R@1 \\
\midrule
\textbf{GeoMoE-G} & \textbf{0.811} & 61.13 &
SinGeo & 16.777 & 59.27 \\
\textbf{GeoMoE-B} & \textbf{0.885} & \textbf{62.39} &
EP-BEV & 33.554 & 58.82 \\
SAIG-S & 6.291 & 36.69 &
VimGeo & 157.286 & 27.74 \\
TransGeo & 16.384 & 39.60 &
Just Zoom In & 1789.49 & 35.13 \\
Sample4Geo & 16.777 & 52.14 &
& & \\
\bottomrule
\end{tabular*}
}
\caption{Query-time accuracy--efficiency on VIGOR-M. DMC is reported in
MMAC/query; bold denotes the two lowest costs and the best R@1 (\%).}
\label{tab:efficiency}
\end{table}

\section{Experiments}

We evaluate accuracy, efficiency, and cross-resolution generalization on two
hierarchical CVGL benchmarks, followed by component and MoE ablations.

\subsection{Experimental Setup}
\label{sec:experimental_setup}

\paragraph{Datasets.}

Just Zoom In contains 278,607 training and 30,956 validation ground-view images, with
69,904 satellite images in a four-level hierarchy. VIGOR-M contains 37,895
training and 37,789 test panoramas across four cities, distributed approximately
evenly, and 17,472 satellite tiles at three levels. Both datasets partition each
parent into a \(4\times4\) grid; VIGOR-M contains 64 tiles at L1.

\paragraph{Evaluation Metrics.}

Following Just Zoom In, \(\mathrm{R}@\tau\mathrm{m}\) is the percentage of
queries whose top-ranked terminal tile center \(\widehat{\mathbf{x}}_m\) lies
within \(\tau\) meters of the ground truth \(\mathbf{x}_m^{*}\):
\begin{equation}
\mathrm{R}@\tau\mathrm{m}
=\frac{100\%}{M}\sum_{m=1}^{M}
\mathbb{I}\!\left[d(\widehat{\mathbf{x}}_m,\mathbf{x}_m^{*})\leq\tau\right].
\end{equation}
We also report the median of
\(d(\widehat{\mathbf{x}}_m,\mathbf{x}_m^{*})\) over all queries; it is less
sensitive to large-error outliers, and lower values are better.

\paragraph{Implementation Details.}

The GeoMoE encoder is initialized from DINOv2 ViT-B/14 \cite{DINOv2}, based on
the Vision Transformer architecture \cite{ViT}; each expert inherits the
corresponding pretrained FFN. The image-level router uses the \(\mathrm{CLS}\)
token and Top-2 routing. We train for 60 epochs on one NVIDIA H200 GPU using
AdamW, a batch size of 128, an initial learning rate of \(10^{-4}\), and
\(\lambda_{\mathrm{aux}}=0.01\). Beam retrieval uses \(K_p=4\).

\paragraph{Cross-Resolution Protocol.}

Following robust CVGL evaluations that transfer conventionally trained models
to altered query conditions \cite{ConGeo,SinGeo}, we evaluate each baseline's
standard L3-trained model on all six galleries without scale-specific
retraining or adaptation. The dense MLJ control uses the same backbone,
initialization, L1/L2/L3 training pairs, and level-wise retrieval objective as
GeoMoE, but retains dense FFNs. L1.5, L2.5, and L3.5 are excluded from joint
training. Standard-baseline comparisons assess robustness to resolution shifts,
whereas MLJ (Dense) serves as a matched control for sparse expert routing.

\begin{table}[!t]
\centering
{\small
\renewcommand{\arraystretch}{0.98}
\setlength{\tabcolsep}{1.5pt}
\begin{tabular*}{\columnwidth}{@{\extracolsep{\fill}}clccc@{}}
\toprule
ID & Setting & R@1 (\%) $\uparrow$ & R@100m (\%) $\uparrow$
& Median (m) $\downarrow$ \\
\midrule
A & 3$\times$ Sample4Geo
& 50.24 & 62.15 & 63.89 \\
B & MLJ
& 57.76\,{\scriptsize $(+7.52)$}
& 72.98\,{\scriptsize $(+10.83)$}
& 55.42\,{\scriptsize $(\downarrow 8.47)$} \\
C & B+MoE
& 61.13\,{\scriptsize $(+3.37)$}
& 76.45\,{\scriptsize $(+3.47)$}
& 53.17\,{\scriptsize $(\downarrow 2.25)$} \\
\textbf{D} & C+Beam+PRC
& \textbf{62.39}\,{\scriptsize $(+1.26)$}
& \textbf{77.82}\,{\scriptsize $(+1.37)$}
& \textbf{52.33}\,{\scriptsize $(\downarrow 0.84)$} \\
\bottomrule
\end{tabular*}
}
\caption{Incremental module ablation on VIGOR-M. A uses three independent
Sample4Geo models; B introduces MLJ, C adds sparse MoE, and D adds Beam+PRC.
Rows A--C use greedy retrieval ($K=1$), and D uses $K=4$. Parentheses report
changes from the preceding row in percentage points or meters.}
\label{tab:module_ablation}
\end{table}

\begin{table}[!t]
\centering
{\small
\renewcommand{\arraystretch}{0.96}
\setlength{\tabcolsep}{3.0pt}
\begin{tabular*}{\columnwidth}{@{\extracolsep{\fill}}cc@{\hspace{8pt}}cc@{}}
\toprule
\multicolumn{2}{c}{Expert-count sweep (Greedy)} &
\multicolumn{2}{c}{MoE-depth sweep (Greedy)} \\
\cmidrule(r){1-2}\cmidrule(l){3-4}
Experts & R@1 (\%) $\uparrow$ & Layers (blocks) & R@1 (\%) $\uparrow$ \\
\midrule
2 & 60.27 & 5 (7--11) & 60.35 \\
3 & 60.93 & 4 (8--11) & 60.94 \\
4 & 61.09 & 3 (9--11) & 60.79 \\
\textbf{5} & \textbf{61.13} & \textbf{2 (10--11)} & \textbf{61.33} \\
6 & 60.76 & 1 (11) & 61.13 \\
7 & 58.92 & 0 (No MoE) & 57.76 \\
\bottomrule
\end{tabular*}
}
\caption{Expert-count and MoE-depth ablation on VIGOR-M using greedy retrieval,
Top-2 routing, and epoch-60 checkpoints. The two sweeps fix one Block-11 MoE
layer and four experts, respectively; blocks are indexed from 0 to 11. Bold
marks each optimum.}
\label{tab:moe_ablation}
\end{table}

\subsection{Comparison with State-of-the-Art Methods}
\label{sec:sota_comparison}

Table~\ref{tab:vigor_m_results} shows that GeoMoE achieves the best recall and
lowest median error on both benchmarks. Because VIGOR-M's L3 tiles cover a
larger ground area, R@100m, R@200m, and R@300m provide more discriminative
thresholds. On Just Zoom In, GeoMoE exceeds the strongest baseline by 2.77,
2.69, and 2.68 percentage points at R@40m, R@50m, and R@100m, respectively.
On VIGOR-M, it improves R@1 by 3.12 points and R@100m, R@200m, and R@300m by
5.28, 6.36, and 6.16 points, while reducing median error by 1.62 m.

Table~\ref{tab:efficiency} reports query-time descriptor-matching computation
(DMC) in million multiply--accumulate operations per query (MMAC/query) for
query--reference descriptor dot products; PRC arithmetic is reported separately.
GeoMoE-G uses \(K=1\), whereas GeoMoE-B uses \(K=4\) with PRC; both encode each
query once against precomputed satellite descriptors. Conventional baselines
scan one L3 gallery, EP-BEV scans two, and Just Zoom In uses autoregressive
decoding. GeoMoE-B achieves 62.39\% R@1 with 0.885 MMAC/query (5.27\% of a
full L3 scan), whereas GeoMoE-G retains 61.13\% R@1 at 0.811 MMAC/query. The
supplement further reports beam-width and PRC sensitivity, including their
interaction, together with cached-CPU and end-to-end measurements of latency,
throughput, and reference storage.

Figure~\ref{fig:qualitative_results} traces one successful VIGOR-M query through
the hierarchy: the ground-truth branch remains among the top four paths at L1
and L2, and the correct L3 tile is ultimately ranked first. This example shows
how beam retrieval preserves plausible paths until the terminal decision.

\subsection{Cross-Resolution Robustness}
\label{sec:cross_resolution}

Table~\ref{tab:single_resolution_results} first evaluates how models trained
under the conventional L3 protocol respond to changes in satellite resolution.
These baselines were strongest near their native scale but degraded markedly on
coarser galleries. Joint multi-scale supervision substantially reduced this
sensitivity: MLJ (Dense) reached 48.11\%, 59.35\%, 71.43\%, and 45.39\% R@1
from L1 through L2.5, while remaining competitive at L3 and L3.5. This contrast
captures the benefit of the multi-scale training paradigm rather than an
MoE-specific gain.

Under the matched L1/L2/L3 protocol, GeoMoE improved over MLJ (Dense) at every
level by 31.11, 0.51, 7.77, 4.01, 11.34, and 12.39 percentage points. Its mean R@1
increased from 57.07\% to 73.81\% on the three training levels and from 37.66\%
to 43.30\% on the three withheld levels. L1.5 and L2.5 measure interpolation
between observed resolutions, whereas L3.5 requires extrapolation beyond the
finest training level. The lower absolute accuracy at L3.5 therefore identifies
fine-scale extrapolation as the remaining cross-resolution challenge.

% \FloatBarrier
\subsection{Ablation Studies}
\label{sec:ablation}

Table~\ref{tab:module_ablation} isolates MLJ, sparse MoE, and Beam+PRC. Relative
to three independent Sample4Geo models~\shortcite{Sample4Geo}, MLJ improved R@1
and R@100m by 7.52 and 10.83 points and reduced median error by 8.47 m. Sparse
MoE added 3.37 and 3.47 points while reducing median error by 2.25 m; Beam+PRC
then added 1.26 and 1.37 points and reduced it by another 0.84 m. From A to D,
the complete system gained 12.15 points in R@1 and 15.67 points in R@100m,
with an 11.56-m reduction in median error.

The expert-count sweep in Table~\ref{tab:moe_ablation} increased R@1 from
60.27\% with two experts to 61.13\% with five. Performance decreased to 60.76\%
with six experts and 58.92\% with seven, placing the optimum at five under the
fixed one-layer setting.

MoE depth was comparatively stable. A single Block-11 layer improved the
no-MoE result from 57.76\% to 61.13\%, while the best two-layer setting reached
61.33\%; configurations with three to five layers remained between 60.35\% and
60.94\%. Because each additional MoE layer introduces another expert set, we
adopt one layer, which was only 0.20 points below the optimum while avoiding
substantial parameter growth.

% \FloatBarrier

\section{Conclusion}

We introduced GeoMoE, a sparse MoE dual encoder for globally supervised
multi-resolution learning and efficient hierarchical retrieval. On VIGOR-M, it
achieved \(62.39\%\) R@1 with \(0.885\) MMAC/query (\(5.27\%\) of an exhaustive
L3 scan) and outperformed a matched dense MLJ control across all six galleries.
It also reached \(95.78\%\) R@40m on Just Zoom In, while ablations isolated gains
from MLJ, sparse routing, and calibrated beam search. Together, these results
support efficient local indexed search over globally comparable descriptors.
Transfer remained strong at withheld resolutions, although L3.5 exposes
fine-scale extrapolation as the main limitation.

% One reference list covers citations in both the main paper and supplement.
\onecolumn
\begin{multicols}{2}
\ArxivBibliography{aaai2027}
\end{multicols}

\clearpage

% Supplement-only float settings. The main-paper layout above remains unchanged.
\makeatletter
\setlength{\@fptop}{0pt}
\setlength{\@fpsep}{14pt}
\setlength{\@fpbot}{0pt plus 1fil}
\setlength{\@dblfptop}{0pt}
\setlength{\@dblfpsep}{14pt}
\setlength{\@dblfpbot}{0pt plus 1fil}
\makeatother
\renewcommand{\topfraction}{0.95}
\renewcommand{\bottomfraction}{0.95}
\renewcommand{\textfraction}{0.06}
\renewcommand{\floatpagefraction}{0.80}
\renewcommand{\dbltopfraction}{0.95}
\renewcommand{\dblfloatpagefraction}{0.80}
\setlength{\textfloatsep}{12pt plus 2pt minus 2pt}
\setlength{\dbltextfloatsep}{12pt plus 2pt minus 2pt}
\raggedbottom

\setcounter{section}{0}
\setcounter{subsection}{0}
\setcounter{figure}{0}
\setcounter{table}{0}
\setcounter{equation}{0}
\renewcommand{\thefigure}{S\arabic{figure}}
\renewcommand{\thetable}{S\arabic{table}}
\renewcommand{\theequation}{S\arabic{equation}}

\twocolumn[
\begin{center}
  {\LARGE\bfseries Supplementary Material}\par
\end{center}
\vspace{0.75em}
]

% The standalone supplement ends with its own bibliography command. Suppress it
% here because the combined preprint already emitted the unified reference list.
\renewcommand{\bibliography}[1]{}
\section{Additional VIGOR-M Dataset Details}
\label{supp:vigor_m_dataset}

The fixed same-area protocol, inherited from VIGOR \cite{VIGOR}, contains
37,895 training queries and 37,789 test queries with disjoint query identities.
Both partitions cover the same four city regions and share the underlying
satellite map. The protocol therefore evaluates map-specific hierarchical
retrieval and localization rather than transfer to unseen geographic regions.
Unless otherwise stated, all VIGOR-M experiments in this work use the
same-area protocol.

\begin{figure}[!h]
  \centering
  \includegraphics[width=0.83\columnwidth]{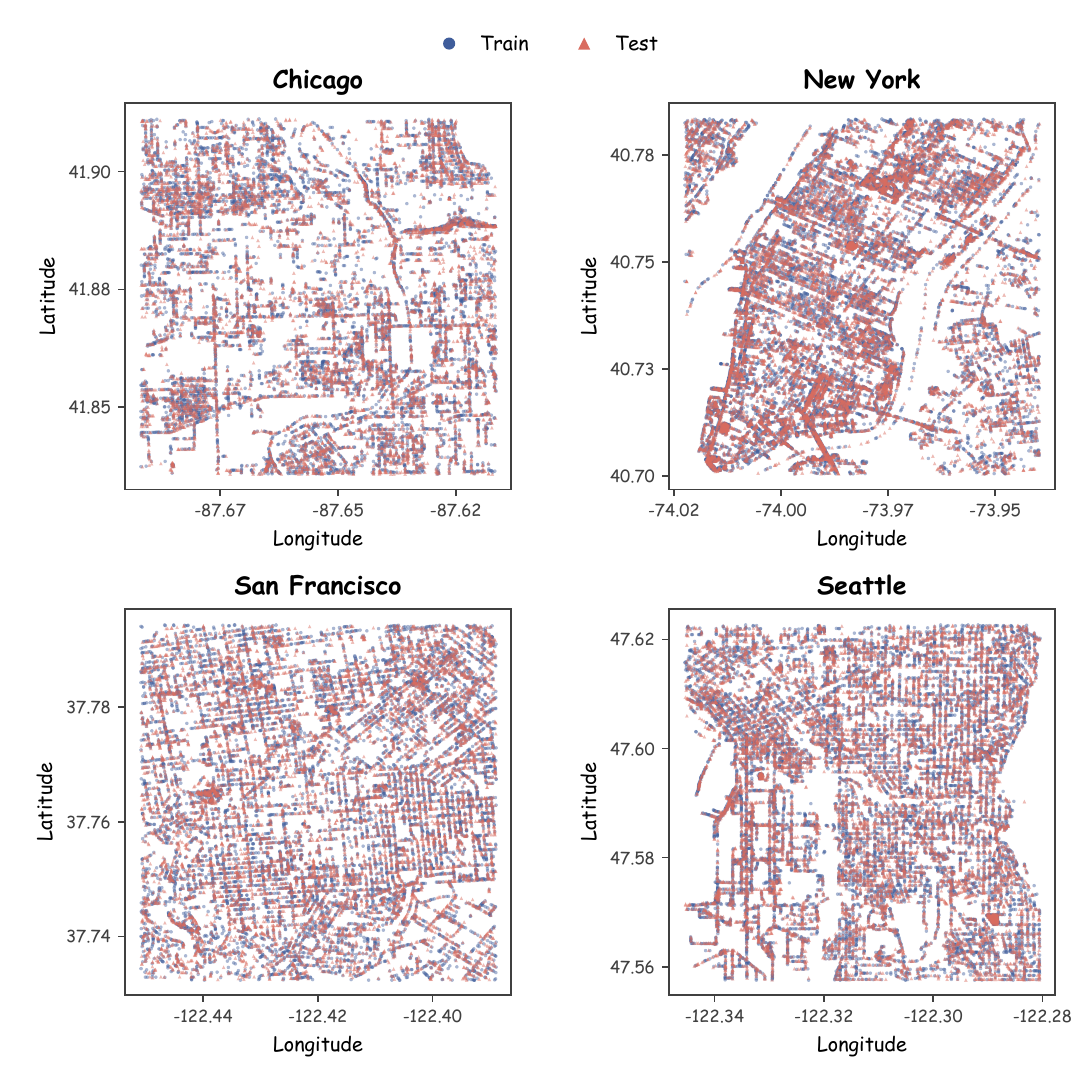}
  \caption{Spatial distribution of all 75,684 VIGOR-M same-area ground queries
  in Chicago, New York, San Francisco, and Seattle. Training (37,895; deep-blue
  circles) and identity-disjoint test queries (37,789; coral triangles) cover
  the same city regions and share the underlying satellite map.}
  \label{supp:fig_vigorm_ground_distribution}
\end{figure}

Figure~\ref{supp:fig_vigorm_ground_distribution} shows all official same-area
queries without spatial subsampling. Train and test identities are disjoint but
spatially interleaved within each city; panel axes are scaled independently for
legibility.

\ifdefined\standalonesupplementbreak
  \standalonesupplementbreak
\fi
\section{Additional Implementation Details}
\label{supp:implementation}

This supplement reports implementation and evaluation details that support the
main paper. The main submission presents the complete method and primary
evidence; here, we specify the training, indexing, and calibration protocols for
reproducibility and provide additional analyses of the learned representation.

\subsection{Shared Encoder and Training Protocol}

The ground and satellite branches share a DINOv2 ViT-B/14 encoder
\cite{DINOv2,ViT}. Before encoding, satellite and ground images are resized to
network inputs of \(384\times384\) and \(432\times768\), respectively. Their
class-token descriptors are \(\ell_2\)-normalized, and each
ground panorama is encoded once for all retrieval levels. Training applies
symmetric InfoNCE \cite{CPC} independently to L1, L2, and L3 pairs, with equal
level weights, temperature \(\tau=0.07\), and label smoothing \(\epsilon=0.1\).
Intermediate levels are withheld.

Within each batch, we exclude repeated identities, satellite labels, and L3
cells. Same-level hard negatives are initialized geographically and refreshed
every four epochs using descriptor similarity \cite{Sample4Geo}. Each
configuration trained in
this work was run once with global seed 1; machine-readable configurations and
scripts accompany the code supplement.

\subsection{Sparse MoE Implementation}

GeoMoE replaces only the Block-11 feed-forward network (FFN) with five experts
initialized from the pretrained FFN; all earlier blocks, attention layers, and
the output head remain shared. The normalized class token selects two experts
per image \cite{SparselyGatedMoE,VMoE}, and their probabilities are renormalized
before the outputs are combined. All image tokens share the same selected
expert route. Because the
router receives no hierarchy-level identifier, routing remains content-adaptive
rather than explicitly level-conditioned. Only the selected experts are
executed, with load balancing weighted by \(\lambda_{\mathrm{aux}}=0.01\).

\subsection{Offline Index and Hierarchical Retrieval}

Each satellite tile is encoded once and indexed with its hierarchy level and
geographic extent. The index combines 1,024 overlapping dense L1 proposal roots
with the L2 and L3 gallery nodes across four cities. The roots form a
\(16\times16\) grid per city at stride fraction 0.25. Retrieval scans all roots
without an oracle city or ground-truth region.

Each retained root \(u\) expands to the nearest boundary-clipped \(4\times4\)
L2 proposal set \(\mathcal{C}_2(u)\), which may overlap those of other roots.
With \(K_1=K_2=4\), vectorized search evaluates all \(4\times16=64\) L2 path
slots and applies parent-local softmax before merging duplicate node IDs by the
maximum accumulated path score. Thus, an L2 node reached from multiple roots is
descriptor-matched once per incoming path slot. The four highest-scoring
distinct L2 nodes then expand through unique native L2--L3 links, yielding 64
terminal candidates. Conditional log-probabilities use \(\tau=0.07\) and are
accumulated along each path before PRC. The protocol uses neither a city oracle
nor test-time adaptation.

DMC counts executed descriptor dot products rather than distinct node IDs. For
768-dimensional descriptors, Beam-4 requires
\((1024+4\times16+4\times16)\times768=0.884736\) MMAC/query. PRC separately
requires \(64(16\times48+48\times48+48)=0.199680\) MMAC/query, bringing the
combined descriptor-matching and calibration cost to 1.084416 MMAC/query.

\subsection{Path Residual Calibration}

Path Residual Calibration (PRC) represents each complete candidate path \(z\)
with the 16-dimensional descriptor in Table~\ref{supp:prc_features}. Statistics
estimated from candidates on the official training split standardize its
features:
\(\widehat{\boldsymbol{\phi}}(z)=
(\boldsymbol{\phi}(z)-\boldsymbol{\mu})/\boldsymbol{\sigma}\).

\begin{table}[!t]
  \centering
  \footnotesize
  \renewcommand{\arraystretch}{1.08}
  \setlength{\tabcolsep}{2pt}
  \begin{tabular*}{\columnwidth}{@{\extracolsep{\fill}}cp{0.38\columnwidth}cp{0.38\columnwidth}@{}}
    \toprule
    ID & Feature & ID & Feature \\
    \midrule
    1 & Total path log-probability
      & 9 & Normalized L2 rank \\
    2 & L1 log-probability
      & 10 & Normalized local L3 rank \\
    3 & Local L2 log-probability
      & 11 & Normalized root entropy \\
    4 & Local L3 log-probability
      & 12 & Root top-1/top-2 margin \\
    5 & L2 raw cosine similarity
      & 13 & Root top-1 probability \\
    6 & L3 raw cosine similarity
      & 14 & Normalized L2 entropy \\
    7 & Cumulative L2 path log-probability
      & 15 & L2 top-1/top-2 margin \\
    8 & Normalized L1 rank
      & 16 & Valid L3-child fraction of the L2 node \\
    \bottomrule
  \end{tabular*}
  \caption{Path features used by PRC. IDs preserve descriptor order within each
  half; normalization statistics are estimated on the official training split.}
  \label{supp:prc_features}
\end{table}

A LayerNorm and two 48-dimensional linear--GELU layers
\cite{LayerNorm,GELU} map the standardized descriptor to a scalar residual.
Zero initialization of the output preserves
the path score at the start of training:
\begin{equation}
S_{\mathrm{cal}}(z)=S_{\mathrm{path}}(z)+
g_{\psi}\!\left(\widehat{\boldsymbol{\phi}}(z)\right).
\end{equation}
After residual addition, \(S_{\mathrm{cal}}\) is a ranking score rather than a
normalized log-probability.

We fit the calibrator only to queries from the official training split for which
the ground-truth L3 tile is present in the generated candidate set. Given
candidate set \(\mathcal{Z}_i\) and positive path \(z_i^+\), the listwise loss is
\begin{equation}
\mathcal{L}_{\mathrm{cal}}=-\frac{1}{|\mathcal{D}_{\mathrm{cov}}|}
\sum_{i\in\mathcal{D}_{\mathrm{cov}}}
\log\frac{\exp(S_{\mathrm{cal}}(z_i^+))}
{\sum_{z\in\mathcal{Z}_i}\exp(S_{\mathrm{cal}}(z))}.
\end{equation}
Training candidates use \(K=3\), whereas inference applies the same calibrator
at \(K=4\). We train for 20 epochs with AdamW \cite{AdamW} (learning rate 0.002,
weight decay 0.0001, batch size 512) and select checkpoints on an internal training
partition. Official test labels are not used for fitting or selection.

\begin{table}[t]
  \centering
  \footnotesize
  \renewcommand{\arraystretch}{1.02}
  \setlength{\tabcolsep}{3.5pt}
  \begin{tabular*}{\columnwidth}{@{\extracolsep{\fill}}ll@{}}
    \toprule
    Item & Configuration \\
    \midrule
    Backbone & DINOv2 ViT-B/14, shared weights \\
    \shortstack[l]{Encoder input\\(ground / satellite)}
      & \(432\times768\) / \(384\times384\) \\
    Training levels & L1, L2, L3 \\
    Batch composition & 128 (43/43/42 by level) \\
    Optimization & AdamW, \(1\times10^{-4}\), cosine decay \\
    Training & 1-epoch warm-up, 60 epochs total \\
    Temperature / smoothing & 0.07 / 0.1 \\
    Augmentation & Synchronized flip, \(p=0.5\); no rotation \\
    Sparse MoE & Block 11, five experts, Top-2 \\
    Hardware & One NVIDIA H200 GPU \\
    \bottomrule
  \end{tabular*}
  \caption{Training configuration of the reported GeoMoE model.}
  \label{supp:training_config}
\end{table}

\section{Additional Evaluation Protocols}
\label{supp:evaluation_protocols}

\subsection{Continuous-Resolution Descriptor Analysis}
\label{supp:continuous_resolution_protocol}

We evaluated a frozen Just Zoom In B11/E5 encoder \cite{justzoomin} at 13
quarter-level scales from L1 to L4. The cohort contains 2,048 distinct
validation locations and 26,624
satellite inputs. PCA was fitted to the unit-normalized final descriptors
without level labels. A linear Ridge probe \cite{RidgeRegression} was trained on native L1--L4 anchors
and evaluated on the nine intermediate levels using five-fold out-of-fold
prediction grouped by 625-m spatial blocks. We used 500 spatial-block bootstrap
replicates \cite{Bootstrap} for uncertainty and decomposed variance into location, level, and
location-by-level components.

\subsection{Matched Scale-Decodability Analysis}
\label{supp:scale_decodability_protocol}

We compared the locked B11/E5 GeoMoE encoder with the original pretrained
DINOv2 ViT-B/14 on identical inputs from 48 geographic centers across four
cities. Each center was rendered at 21 scales from 1.0 to 3.0 in increments of
0.1, yielding 1,008 inputs per encoder. Separate Ridge probes
(\(\alpha=0.2\)) were fitted to the unit-normalized final descriptors using
alternating 0.2-level anchors and evaluated on the ten intervening scales. We
used identical four-fold GroupKFold partitions by geographic center and
estimated paired uncertainty from 10,000 center-level bootstrap replicates.
No center or scale was excluded.

\subsection{Unsupervised Feature-Map Analysis}
\label{supp:semantic_protocol}

For each L3 tile, we summarized the final post-normalization spatial tokens by
their channel-wise mean and standard deviation over the \(27\times27\) patch
grid, producing a normalized 1,536-dimensional descriptor. Geographic
coordinates and RGB statistics were not used. We retained 64 principal
components and evaluated K-means solutions \cite{KMeans} for
\(K=2,\ldots,12\). The cluster count was selected using standard
internal-validity metrics, including the silhouette coefficient
\cite{Silhouette}, subject to a minimum pairwise ARI of 0.90
\cite{AdjustedRand} across five seeds; the final fit used 20
initializations. The analysis included 3,845 valid Just Zoom In cells and a pooled
set of 16,384 VIGOR-M cells from four cities without city labels. For external
validation, we compared the Just Zoom In clusters with the 2024 District of
Columbia Existing Land Use layer on 1,903 high-confidence cells. Confidence
intervals for NMI, ARI, purity, and V-measure \cite{AdjustedRand,VMeasure} were
estimated from 1,000 bootstrap replicates over \(8\times8\) spatial blocks.

\section{Additional Results and Analyses}
\label{supp:additional_results}

\subsection{Search Accuracy and Efficiency}

Table~\ref{supp:search_variants} separates fixed beam scores from residual
calibration. Calibration improves R@1 at both beam widths without changing the
number of terminal candidates. We select the official \(K=4\) configuration
because it yields the highest R@1 without test-time adaptation. Flat retrieval
remains an exhaustive reference and shows that hierarchical search is not
lossless.

\begin{table*}[!t]
  \centering
  \small
  \renewcommand{\arraystretch}{1.05}
  \setlength{\tabcolsep}{6pt}
  \begin{tabular}{lrrrr}
    \toprule
    Retrieval variant & R@1 (\%) & R@5 (\%) & R@100m (\%)
      & Terminal candidates \\
    \midrule
    Flat exhaustive L3 & 63.0078 & 82.1324 & 78.2318 & Full L3 gallery \\
    Fixed beam, \(K=3\) & 62.2139 & 81.4073 & 77.6760 & 35.61 \\
    Fixed beam, \(K=4\) & 62.2297 & 81.6402 & 77.6787 & 47.00 \\
    Calibrated beam, \(K=3\) & 62.3726 & 81.7487 & \textbf{77.8189} & 35.61 \\
    \textbf{Calibrated beam, \(K=4\)} & \textbf{62.3912}
      & \textbf{81.9868} & 77.8163 & 47.00 \\
    \bottomrule
  \end{tabular}
  \caption{VIGOR-M search variants evaluated on all 37,789 official test
  queries under the inherited VIGOR same-area protocol \cite{VIGOR}. Bold
  indicates the best hierarchical result in each recall column.}
  \label{supp:search_variants}
\end{table*}

GeoMoE-G (greedy) reaches \(61.13\%\) R@1 on VIGOR-M with DMC of 0.811
MMAC/query. GeoMoE-B (Beam-4 with PRC) reaches \(62.39\%\) R@1 with 0.885
MMAC/query, or \(5.27\%\) of the 16.777 MMAC/query required by the full L3 scans
of Sample4Geo and SinGeo. DMC
excludes PRC arithmetic, data movement, indexing overhead, and hardware
utilization.

\FloatBarrier
\subsection{Beam-Width and PRC Sensitivity}
\label{supp:beam_prc_sensitivity}

\begin{figure*}[!t]
  \centering
  \includegraphics[width=\textwidth]{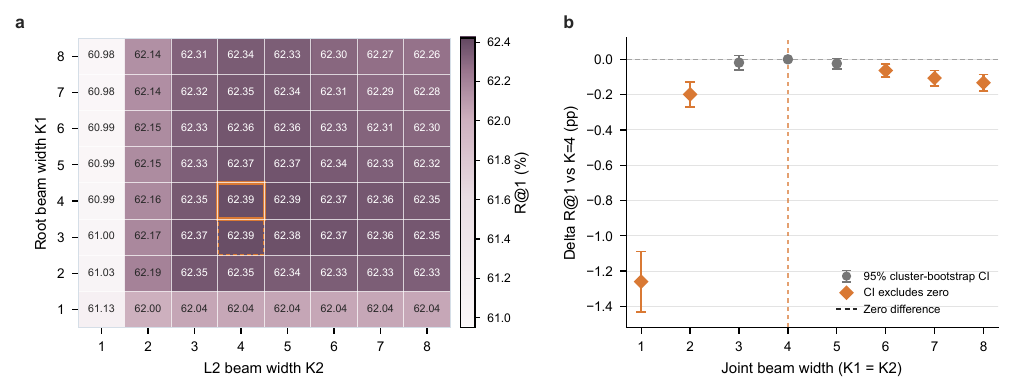}
  \caption{Beam-width sensitivity on VIGOR-M. \textbf{a}, R@1 over all 64
  combinations of root width \(K_1\) and L2 width \(K_2\). The solid outline
  marks the retained \((4,4)\) setting, and the dashed outline marks the tied
  asymmetric maximum at \((3,4)\). \textbf{b}, Paired R@1 differences along
  \(K_1=K_2=K\) relative to \(K=4\). Points show full-test estimates; error bars
  are two-sided 95\% geographic-cluster bootstrap intervals, and diamonds mark
  intervals excluding zero. The analysis uses 37,789 queries, 998 L2 clusters,
  and 5,000 bootstrap replicates. The checkpoint and training-fitted PRC remain
  fixed across widths.}
  \label{supp:fig_vigorm_beam_sensitivity}
\end{figure*}

These experiments address two related questions: whether retrieval is sensitive
to the widths used at successive hierarchy levels, and whether the benefit of
PRC depends on the diversity of the retained candidates. On VIGOR-M, we evaluate
all 64 combinations of root width \(K_1\) and L2 width \(K_2\) on the complete
test set. The B11/E5 checkpoint is fixed, and one PRC fitted on
\((K_1,K_2)=(3,3)\) candidate paths from the official training split is applied
unchanged to every width. No width-specific fitting or test-set adaptation is
used. Along \(K_1=K_2=K\), R@1 rises from \(61.132\%\) at \(K=1\) to
\(62.391\%\) at \(K=4\), while \(K=3\)--5 forms a stable plateau
(Fig.~\ref{supp:fig_vigorm_beam_sensitivity}). Relative to \(K=4\), the R@1
differences at \(K=3\) and \(K=5\) are \(-0.019\) pp (95\% CI,
\(-0.060\)--0.021) and \(-0.024\) pp (95\% CI, \(-0.053\)--0.005), respectively.
The interval first excludes zero at \(K=6\), where R@1 is lower by 0.064 pp.

The two beam dimensions are not equally influential. Most of the grid-level
gain arises as \(K_2\) increases to 3--4, whereas retaining more than three roots
provides little additional benefit. The full grid reaches the same maximum at
\((K_1,K_2)=(3,4)\) and \((4,4)\); their aggregate R@1 values are identical,
although each configuration corrects 12 queries that the other misses. The
asymmetric setting reduces DMC by only 1.39\%, so we retain \((4,4)\) to avoid a
second independently tuned width. This choice is also consistent across cities:
the diagonal optimum occurs at \(K=3\) for Chicago and New York and first occurs
at \(K=4\) for San Francisco and Seattle, with no city benefiting from \(K>4\).
Beyond the
plateau, computation continues to grow: increasing \(K\) from 4 to 8 nearly
doubles the mean final L3 candidate count from 47.00 to 92.43 and raises
descriptor-plus-PRC arithmetic from 1.084 to 1.382 MMAC/query, while reducing
R@1 by 0.132 pp. Wider search therefore introduces additional paths that the
fixed scoring function cannot rank more accurately.

A complementary Just Zoom In analysis isolates beam search by disabling PRC for
\(K=1,\ldots,8\) on all 30,956 validation queries. Increasing \(K\) from 1 to 4
raises ground-truth L4 coverage from \(90.205\%\) to \(98.385\%\) and improves
R@1 by 1.040 pp, while descriptor DMC increases by 19.0\%, from 0.188 to 0.223
MMAC/query. Most of the gain is already obtained at \(K=2\), which reaches
\(97.232\%\) coverage and improves R@1 by 1.021 pp. Increasing \(K\) from 4 to
8 further raises coverage to \(98.776\%\) and the mean final candidate count
from 25.12 to 46.88, but leaves R@1 effectively unchanged. Candidate
reachability is therefore necessary but not sufficient: once the correct path
is retained, additional low-scoring alternatives do not improve the final
decision under the uncalibrated path score.

\begin{figure*}[!t]
  \centering
  \includegraphics[width=\textwidth]{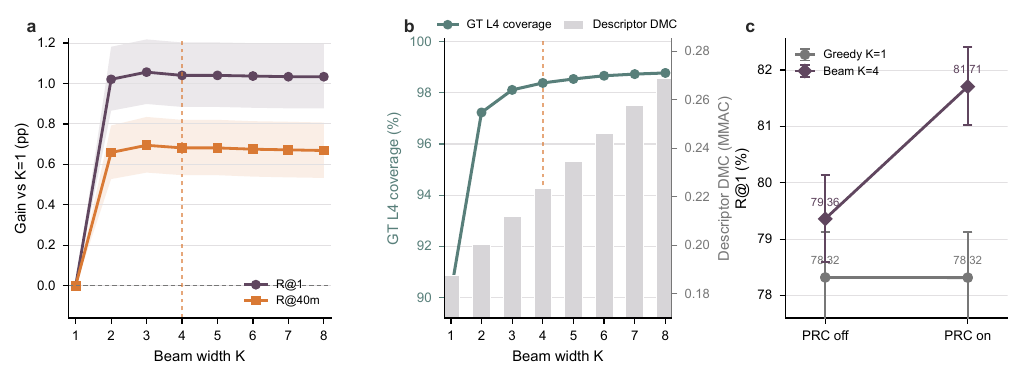}
  \caption{Beam-width and PRC sensitivity on Just Zoom In. \textbf{a}, R@1 and
  R@40m gains over greedy \(K=1\) for path-score-only beam widths
  \(K=1,\ldots,8\); shaded bands denote two-sided 95\% dense-L2 cluster-bootstrap
  intervals. \textbf{b}, Ground-truth L4 candidate coverage and descriptor DMC.
  \textbf{c}, R@1 under the \(K\in\{1,4\}\) by PRC-off/on factorial design. PRC
  is fitted separately at each width using only the official training split.
  All panels use 30,956 validation queries, 2,361 spatial clusters, 5,000
  bootstrap replicates, and the same locked B11/E5 checkpoint.}
  \label{supp:fig_justzoomin_beam_prc}
\end{figure*}

We next evaluate a \(K\in\{1,4\}\) by PRC-off/on factorial design, fitting a
separate PRC on the official training split for each candidate distribution.
At \(K=1\), PRC produces eight R@1 repairs and nine regressions, corresponding
to \(-0.003\) pp (95\% CI, \(-0.032\)--0.025; exact \(p=1.000\)). At \(K=4\),
it produces 1,120 repairs and 394 regressions, improving R@1 by 2.345 pp (95\%
CI, 1.990--2.715) and R@40m by 1.509 pp. The resulting Beam-by-PRC interaction
is positive for both R@1 (2.348 pp; 95\% CI, 1.991--2.720) and R@40m (1.505 pp;
95\% CI, 1.211--1.807). The complete \(K=4\) + PRC system consequently improves
R@1 from \(78.321\%\) to \(81.706\%\), R@40m from \(93.591\%\) to \(95.781\%\),
and median error from 16.692 to 16.432 m relative to greedy search
(Fig.~\ref{supp:fig_justzoomin_beam_prc}). The interaction clarifies the division
of labor: beam search prevents irreversible path pruning, whereas PRC becomes
effective only when multiple plausible branches are available to re-rank.

The two datasets provide complementary rather than numerically interchangeable
tests. VIGOR-M uses one PRC transferred unchanged across the width sweep,
whereas the Just Zoom In width curve disables PRC and the factorial comparison
fits a calibrator for each candidate distribution. Their hierarchy depths,
gallery structures, and distance-recall thresholds also differ. Absolute DMC
and effect sizes should therefore be interpreted within each dataset. The shared
result is the operating regime: moderate widths recover most reachable paths,
while wider search increases candidate and matching costs without consistent
accuracy gains. We consequently retain \(K=4\) as a stable cross-dataset
operating point rather than a uniquely optimal width. The Just Zoom In intervals
use 5,000 resamples of 2,361 dense-L2 spatial clusters and quantify geographic
sampling uncertainty for one fixed checkpoint, not training-seed variability.

\subsection{Deployment-Oriented Efficiency}
\label{supp:deployment_efficiency}

DMC isolates query--gallery dot products but excludes index access,
top-\(K\) selection, hierarchy control, and PRC. We therefore evaluate two
complementary execution boundaries using the same fixed request trace. The
cached-retrieval protocol replays 100,000 requests on one CPU thread, with query
descriptors and gallery state resident in memory. Each run follows 1,000 warm-up
requests. The end-to-end protocol uses the first 10,000 requests after 2,000
warm-up requests. It measures the complete path from in-memory JPEG bytes to the
final retrieval decision on one NVIDIA H200 at batch size one. Each method runs
in a separate process, with OpenCV and PyTorch restricted to one CPU thread.

\begin{figure}[!t]
  \centering
  \includegraphics[width=\columnwidth]{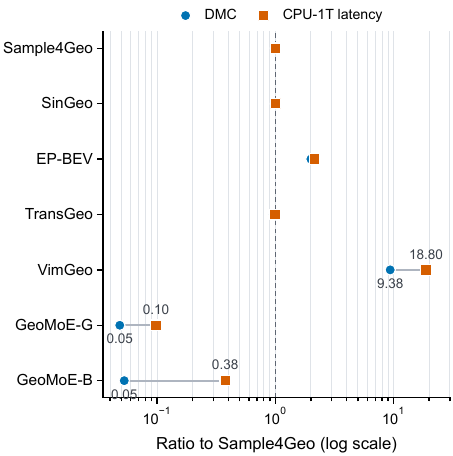}
  \caption{Analytical DMC and measured single-threaded CPU cached-retrieval
  latency on VIGOR-M, both normalized to Sample4Geo. All methods replay the same
  100,000 requests after identical warm-up; lower is better. The difference
  between the two ratios reflects retrieval overhead excluded from DMC.
  Baselines follow their published formulations
  \cite{Sample4Geo,SinGeo,PanoramaBEV,TransGeo,VimGeo}.}
  \label{supp:fig_cpu_efficiency}
\end{figure}

Figure~\ref{supp:fig_cpu_efficiency} shows that GeoMoE-G and GeoMoE-B
reduce DMC to approximately \(0.05\times\) that of Sample4Geo. Their measured
CPU latency ratios are \(0.10\times\) and \(0.38\times\), corresponding to
\(10.22\times\) and \(2.66\times\) speedups, respectively. The smaller
wall-clock gains account for top-\(K\) selection, indexed gathers, hierarchy
control, and PRC, which are not included in DMC.

\begin{figure*}[!t]
  \centering
  \includegraphics[width=\textwidth]{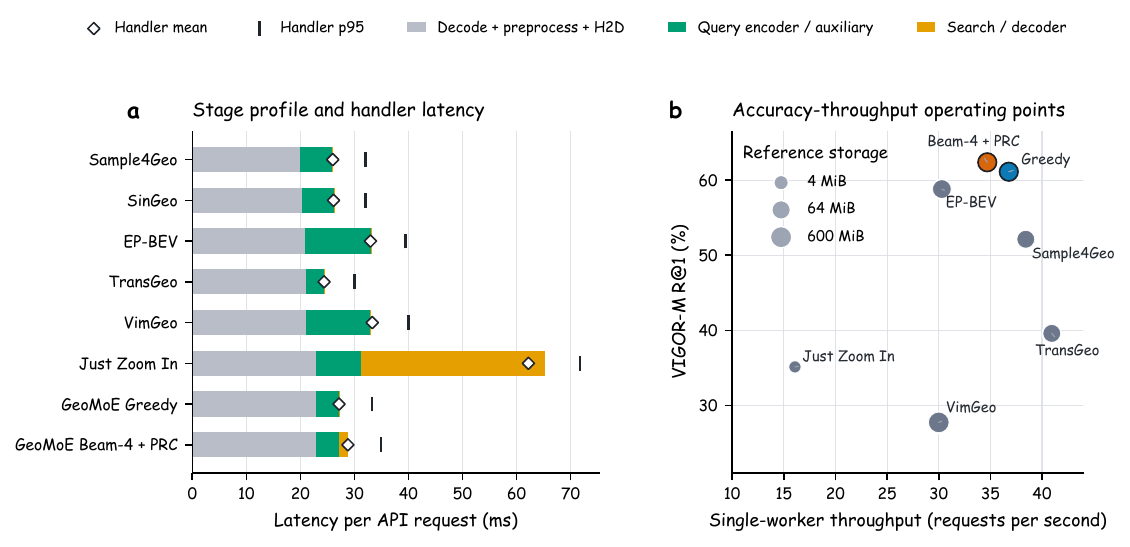}
  \caption{Deployment-oriented VIGOR-M efficiency under a shared
  batch-size-one H200 protocol. (a) Stage-wise latency with independently
  measured handler mean and p95. (b) Full-test-set R@1 versus measured
  single-worker throughput; marker area denotes reference storage.}
  \label{supp:fig_api_efficiency}
\end{figure*}

Figure~\ref{supp:fig_api_efficiency} summarizes the end-to-end boundary.
GeoMoE-G and GeoMoE-B require 27.16 and 28.79 ms per request,
respectively, compared with 25.98 ms for Sample4Geo. Their throughputs are 36.8
and 34.7 requests/s, while Sample4Geo reaches 38.4 requests/s. GeoMoE
nevertheless improves R@1 from 52.14\% to 61.13\% and 62.39\%, while reducing
reference storage from 64 to 54 MiB. Image processing and query encoding
dominate the complete handler, so the cached-search speedup does not imply a
proportional end-to-end improvement. These measurements exclude network
transport, request queuing, authentication, and dynamic batching.

\subsection{Continuous Cross-Resolution Encoding}
\label{supp:continuous_resolution_discussion}

GeoMoE is trained on L1, L2, and L3 and obtains \(59.86\%\), \(49.40\%\), and
\(20.63\%\) R@1 on the withheld L1.5, L2.5, and L3.5 levels, respectively.
Fine-scale transfer at L3.5 therefore remains the most difficult regime.

\begin{figure}[!t]
  \centering
  \captionsetup{justification=raggedright,singlelinecheck=false}
  \includegraphics[width=\columnwidth]{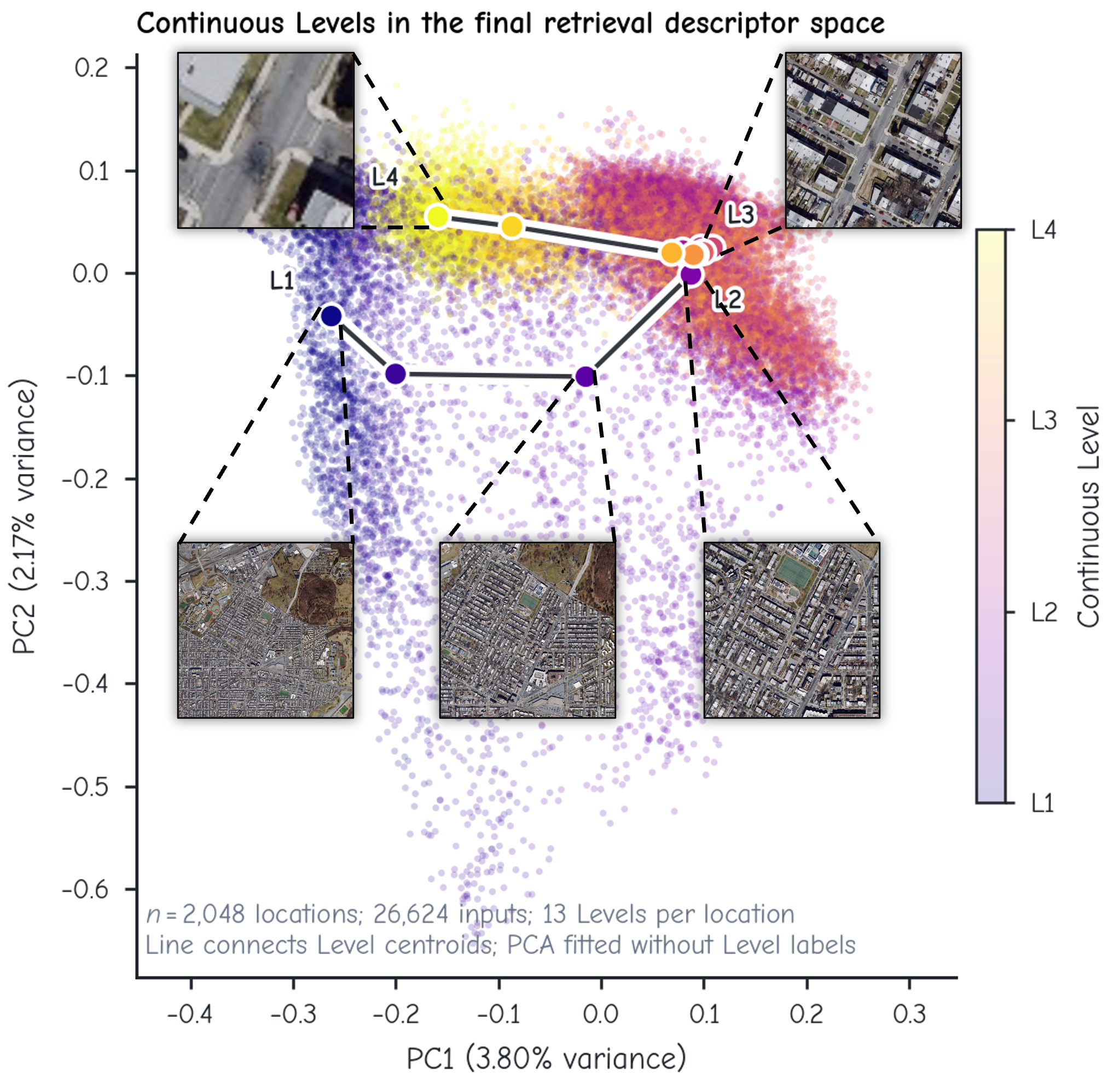}
  \caption{Continuous-resolution structure in the final retrieval descriptor
  space. PCA projects 26,624 unit-normalized descriptors from 2,048 locations
  observed at 13 scales. Points denote individual inputs; outlined points and the
  connecting line show level centroids from L1 to L4, and insets show
  representative native-level inputs. PCA was fitted without level labels. PC1
  and PC2 explain 3.80\% and 2.17\% of total variance, respectively; the
  projection is descriptive and does not preserve all distances in the original
  768-dimensional space.}
  \label{supp:fig_continuous_resolution}
\end{figure}

In the separate Just Zoom In diagnostic, the 13 level centroids trace a continuous
but nonlinear trajectory (Fig.~\ref{supp:fig_continuous_resolution}). A probe
trained only on native L1--L4 anchors predicts intermediate levels with
\(R^2=0.526\), Spearman \(\rho=0.765\), and a mean absolute error of 0.459
levels. However, hierarchy level explains only \(4.37\%\) of descriptor
variance, compared with \(44.40\%\) for location and \(51.24\%\) for their
interaction. These results indicate that resolution is ordered within the
descriptor space but remains strongly content-dependent. They do not establish
a one-to-one mapping between experts and hierarchy levels because routing is
conditioned on image content rather than an explicit level identifier.

We next evaluated complete L1.5--L2.5--native-L3 retrieval on the same 18,797
query IDs and aligned native-L3 ground truth for all three methods
(Table~\ref{supp:intermediate_scale_transfer}).

\begin{table}[t]
  \centering
  \scriptsize
  \renewcommand{\arraystretch}{1.05}
  \setlength{\tabcolsep}{1.5pt}
  \begin{tabular*}{\columnwidth}{@{\extracolsep{\fill}}lrrrrr@{}}
    \toprule
    Method & R@1 & R@100m & R@200m & R@300m & Median \\
    & (\%) & (\%) & (\%) & (\%) & (m) \\
    \midrule
    \textbf{GeoMoE (ours)} & \textbf{49.247} & \textbf{61.536}
      & \textbf{67.899} & \textbf{71.192} & \textbf{65.220} \\
    Sample4Geo & 7.129 & 10.470 & 16.864 & 24.546 & 795.034 \\
    Just Zoom In & 0.112 & 0.309 & 1.261 & 2.750 & 1,364.457 \\
    \bottomrule
  \end{tabular*}
  \caption{Matched intermediate-scale transfer on VIGOR-M. All methods use the
  same 18,797 query IDs and aligned native-L3 ground truth. GeoMoE uses one
  B11/E5 model with training-only PRC; Sample4Geo \cite{Sample4Geo} uses three
  uncalibrated L1/L2/L3 models. Both share galleries and \(K_1=K_2=4\).
  Just Zoom In \cite{justzoomin} uses a native greedy hierarchy, a ground-truth
  L0.5 region and city oracle, and
  L3.5-to-L3 folding. The comparison is method-level because protocols differ.}
  \label{supp:intermediate_scale_transfer}
\end{table}

GeoMoE reached \(49.247\%\) R@1 and \(61.536\%\), \(67.899\%\), and
\(71.192\%\) recall within 100, 200, and 300 m, respectively, with a median
error of 65.220 m. The corresponding R@1 values were \(7.129\%\) for the
three-model Sample4Geo cascade and \(0.112\%\) for Just Zoom In. Without PRC,
the raw GeoMoE Beam-4 R@1 remained \(41.198\%\), exceeding Sample4Geo by
34.069 percentage points. The advantage therefore cannot be attributed to
path-score calibration alone.

The stage diagnostics localize where the methods diverge. GeoMoE retained the
correct path for \(77.103\%\) of queries after L1.5 and \(65.702\%\) after
L2.5, whereas Sample4Geo fell from \(60.988\%\) to \(17.572\%\). Just Zoom In
greedy stage hits decreased from \(6.820\%\) to \(0.351\%\), despite its
ground-truth coarse-region oracle. Together, these results are consistent with
stronger transfer of GeoMoE descriptors to unseen intermediate resolutions,
but they constitute a method-level comparison because the backbones, model
counts, and inference protocols are not architecture-controlled.

Original DINOv2 retained a strongly decodable absolute-scale axis on the unseen
scales: its MAE was \(0.106\),
\(R^2=0.941\), and Spearman \(\rho=0.971\), compared with \(0.363\),
\(0.393\), and \(0.712\) for GeoMoE. The paired MAE difference
(DINOv2 minus GeoMoE) was \(-0.258\) levels (95\% CI,
\([-0.288,-0.228]\)); GeoMoE had lower scale-decoding error at none of the 48
centers.

Within-location ordering showed the same contrast. GeoMoE correctly ordered
\(65.0\%\) of adjacent scale pairs and \(88.1\%\) of all scale pairs, with a
median trajectory Spearman correlation of \(0.929\). Original DINOv2 reached
\(91.4\%\), \(98.7\%\), and \(0.997\), respectively. Thus, GeoMoE preserves
substantial ordinal information along each scale trajectory even though the
absolute scale coordinate is less accessible to a global linear probe.

The difference was robust to the probe regularization. Across the tested Ridge
values from \(\alpha=0.001\) to \(10\), GeoMoE unseen-scale MAE ranged from \(0.354\) to
\(0.477\), whereas DINOv2 ranged from \(0.103\) to \(0.214\) and remained lower
at every setting.

The probe and retrieval results measure complementary properties: the former
asks whether a global linear map recovers absolute scale, whereas the latter
asks whether descriptors remain comparable across unseen resolutions. GeoMoE's
strong retrieval despite lower decodability is consistent with scale-stable
matching, not scale insensitivity or evidence that suppressing scale causes the
gain. Although the main-paper retrieval comparison includes a matched dense MLJ
control, attributing reduced scale decodability to MoE would require applying
the same probe to that control.

\subsection{Semantic Organization of Spatial Feature Maps}
\label{supp:semantic_discussion}

Following SNAP's finding that ground--overhead alignment for positioning can
induce high-level semantics without labels \cite{SNAP}, we test whether GeoMoE's
cross-view supervision yields land-use correspondence and spatial coherence in
satellite features.

\begin{figure*}[!t]
  \centering
  \includegraphics[width=0.95\textwidth]{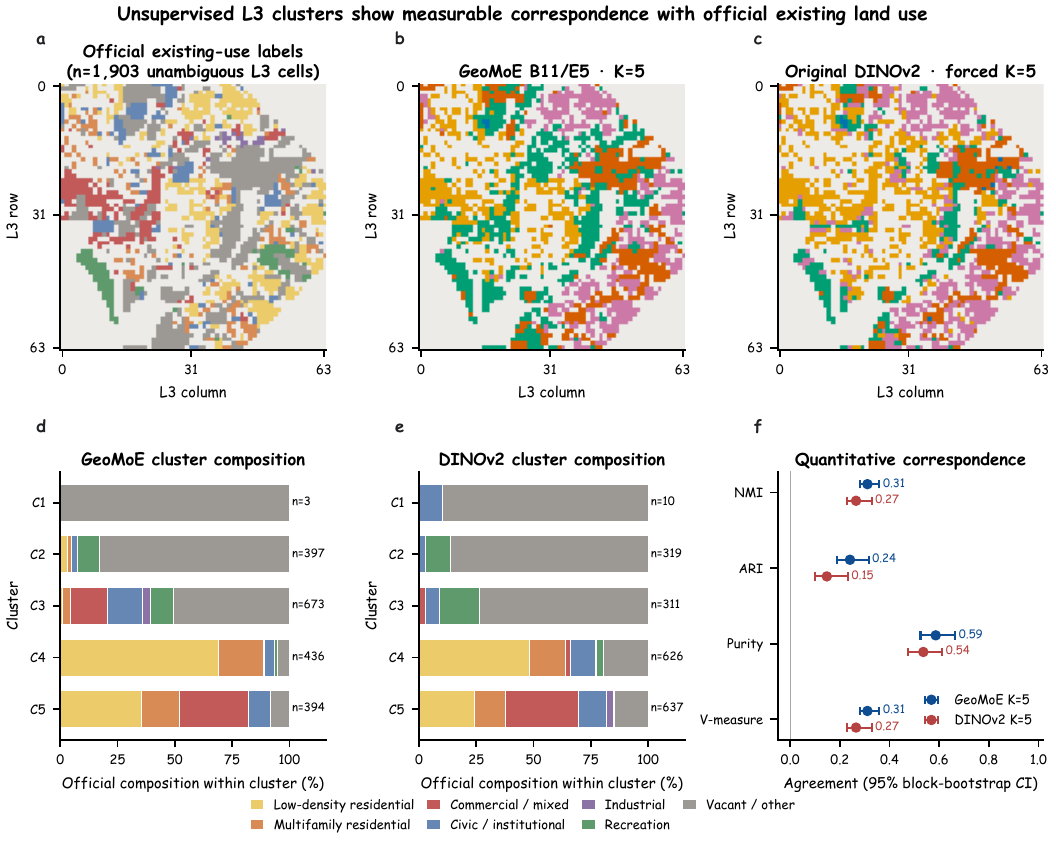}
  \caption{Correspondence between unsupervised L3 clusters and official existing
  land use in Washington, DC. \textbf{a}, The 1,903 high-confidence L3 cells
  derived from the 2024 DC Existing Land Use layer; white cells fail at least one
  jurisdiction, image-quality, official-coverage, or dominant-class criterion.
  \textbf{b,c}, Matched \(K=5\) assignments from GeoMoE B11/E5 and the original
  DINOv2 encoder on the same eligible cells. \textbf{d,e}, Official land-use
  composition within each cluster; labels at bar ends give the number of
  evaluated cells. \textbf{f}, NMI, ARI, purity, and V-measure with 95\%
  intervals from 1,000 spatial block-bootstrap replicates using \(8\times8\)
  L3-cell blocks. GeoMoE yields higher point estimates for all four metrics.}
  \label{supp:fig_landuse_validation}
\end{figure*}

On Just Zoom In, the descriptors form a stable five-cluster partition with measurable
correspondence to an independent land-use reference
(Fig.~\ref{supp:fig_landuse_validation}). At matched \(K=5\), GeoMoE obtains NMI
\(0.311\), ARI \(0.241\), and purity \(0.586\), compared with \(0.265\),
\(0.147\), and \(0.536\), respectively, for the original DINOv2 encoder. The
paired bootstrap intervals lie above zero for all displayed metrics. This
comparison captures the combined effect of fine-tuning and the MoE architecture,
not the causal contribution of either component alone.

\begin{figure*}[!t]
  \centering
  \includegraphics[width=\textwidth]{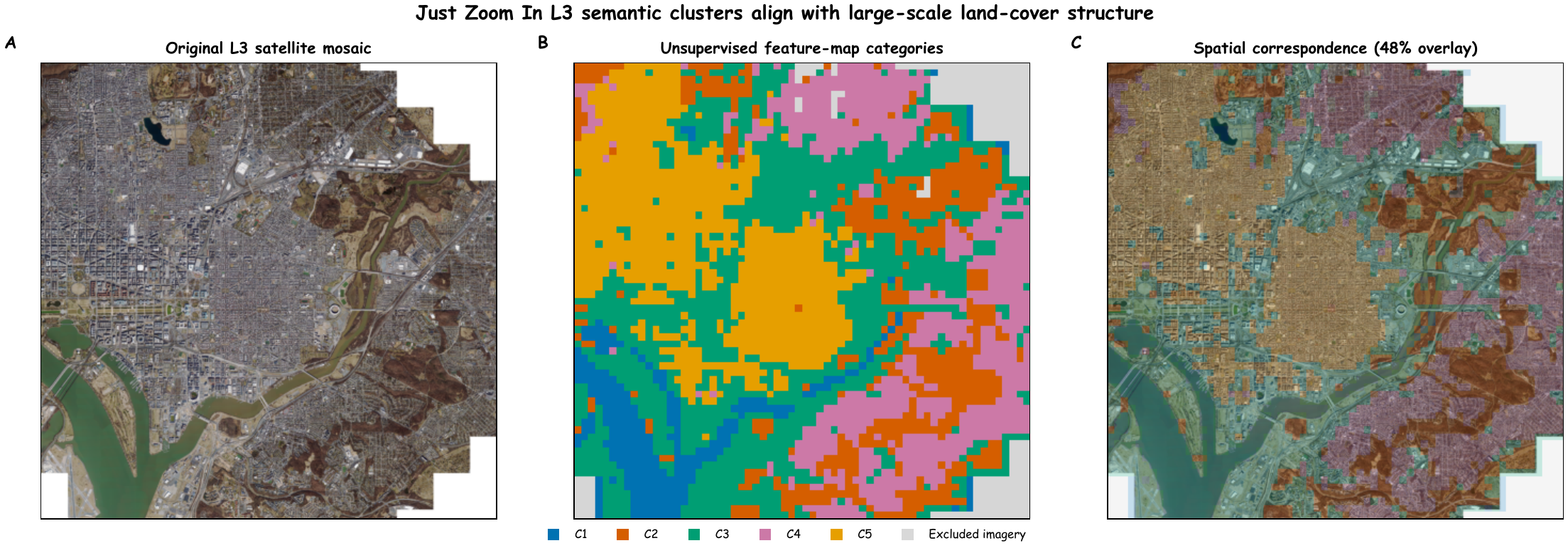}
  \caption{Spatial organization of GeoMoE L3 clusters in Just Zoom In.
  \textbf{A}, Satellite mosaic formed by the complete 4,096-cell native L3 grid.
  \textbf{B}, Five unsupervised clusters obtained from the 3,845 cells that pass
  the prespecified image-quality criteria; gray cells were excluded because of
  boundary black regions or near-uniform imagery. \textbf{C}, Cluster assignments
  overlaid on the satellite mosaic at 48\% opacity. Large contiguous regions
  correspond to coarse visual structures related to water, natural open space,
  roads or large facilities, residential density, and the urban core. Geographic
  coordinates and RGB statistics were not used for PCA, cluster-count selection,
  or K-means fitting.}
  \label{supp:fig_spatial_semantics}
\end{figure*}

Spatial organization provides a complementary test. Four-neighbor cluster
agreement reaches \(0.742\), compared with \(0.236\)
after label-preserving spatial permutation (\(p=0.001\)); see
Fig.~\ref{supp:fig_spatial_semantics}. The partition therefore reflects regional
image structure rather than cluster proportions alone. Nevertheless, post hoc
labels such as water, open space, infrastructure, and urban core describe visual
tendencies rather than supervised functional zones.

\begin{figure*}[!t]
  \centering
  \includegraphics[width=\textwidth]{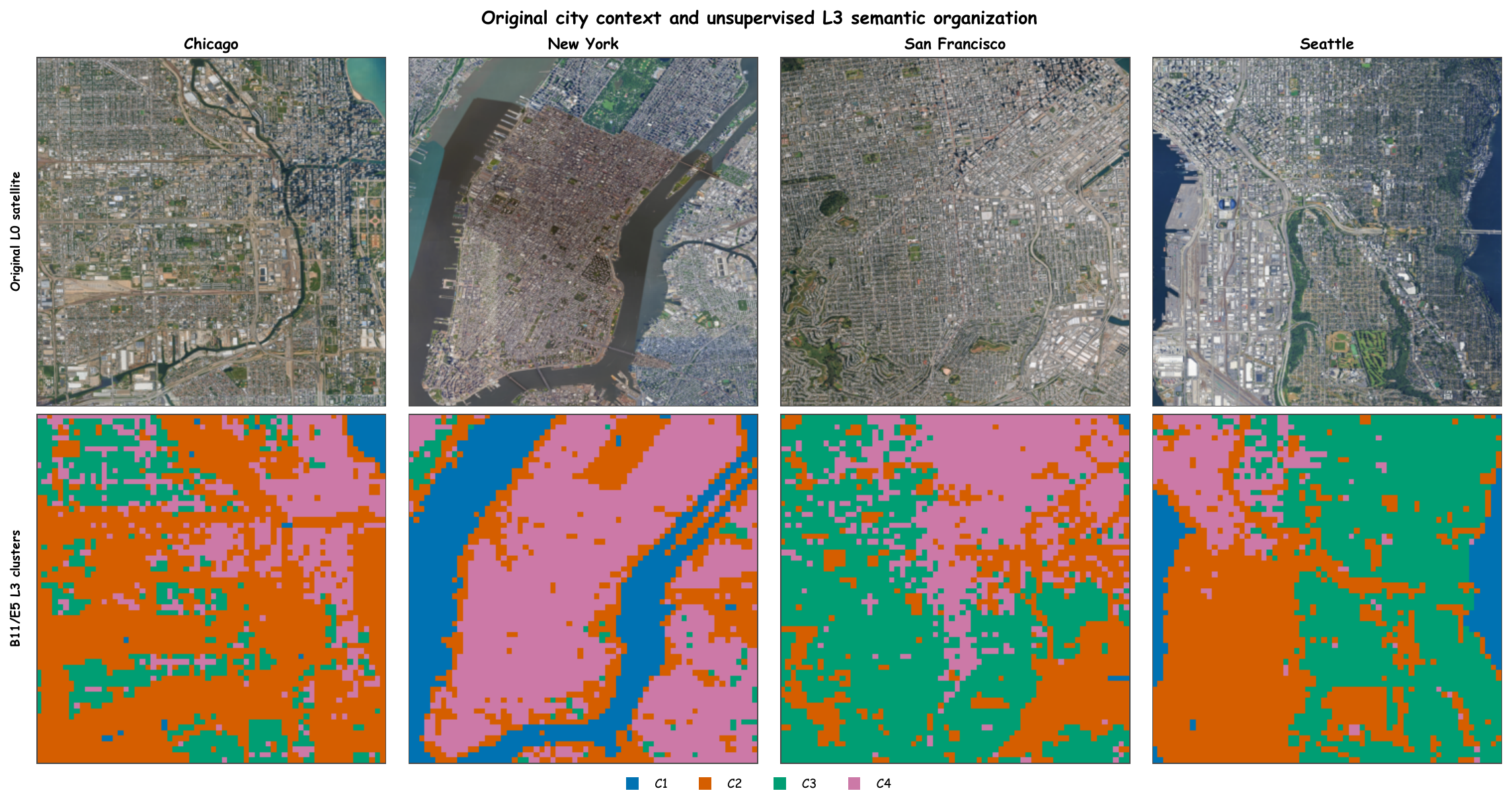}
  \caption{Original VIGOR-M L0 scenes and shared L3 semantic components across
  four cities. The top row shows the L0 satellite images for Chicago, New York,
  San Francisco, and Seattle; the bottom row shows the corresponding
  \(64\times64\) L3 cluster maps. All 16,384 descriptors were pooled for a single
  PCA--K-means fit, so C1--C4 are shared cross-city components rather than
  city-specific clustering solutions. Post hoc inspection associates C1--C4
  with open water, bright impervious infrastructure, vegetated low-density
  residential texture, and dense urban cores, respectively. Differences between
  cities are expressed primarily through the proportions and spatial arrangements
  of these shared components.}
  \label{supp:fig_cross_city_semantics}
\end{figure*}

Pooled VIGOR-M clustering provides a third test and yields four shared components rather than city-specific
groups (Fig.~\ref{supp:fig_cross_city_semantics}). The solution is stable (mean
ARI \(0.998\)) but weakly separated (silhouette \(0.117\)). Low city--cluster
NMI (\(0.158\)) contrasts with \(93.46\%\pm0.46\%\) linear-probe accuracy,
indicating that city identity is encoded through component mixtures and
continuous shifts rather than one cluster per city. Overall, the feature maps
contain spatially coherent components of coarse urban morphology. The post hoc
clusters are neither supervised functional zones nor an unsupervised taxonomy
of city identity.

\subsection{Hierarchical Failure Modes}
\label{supp:failure_modes}

We assigned each exact L3 error to the first stage at which the ground-truth
path was lost under the locked B11/E5 Top-2 MoE, fixed Beam-4, and PRC protocol.
Among 14,212 failures from 37,789 test queries, 993 (6.99\%) were L1
reachability failures, 2,152 (15.14\%) resulted from L2 pruning, and 11,067
(77.87\%) were L3 mis-rankings. Their median geographic errors were 3.37 km,
0.89 km, and 99 m, respectively. Early search failures were less frequent but
more geographically consequential, whereas most residual errors arose from
local fine-grained ranking.

This stage profile separates two optimization targets. L1 and L2 losses account
for \(22.13\%\) of exact failures, but their larger median distances make them
important to the long-range error tail. The dominant L3 category instead points
to local descriptor discrimination, terminal-candidate diversity, and residual
calibration as the main opportunities for improving exact retrieval. These
priorities are diagnostic rather than causal because the categories are defined
conditionally on the current search policy.

\begin{figure*}[!t]
  \centering
  \includegraphics[width=0.82\textwidth]{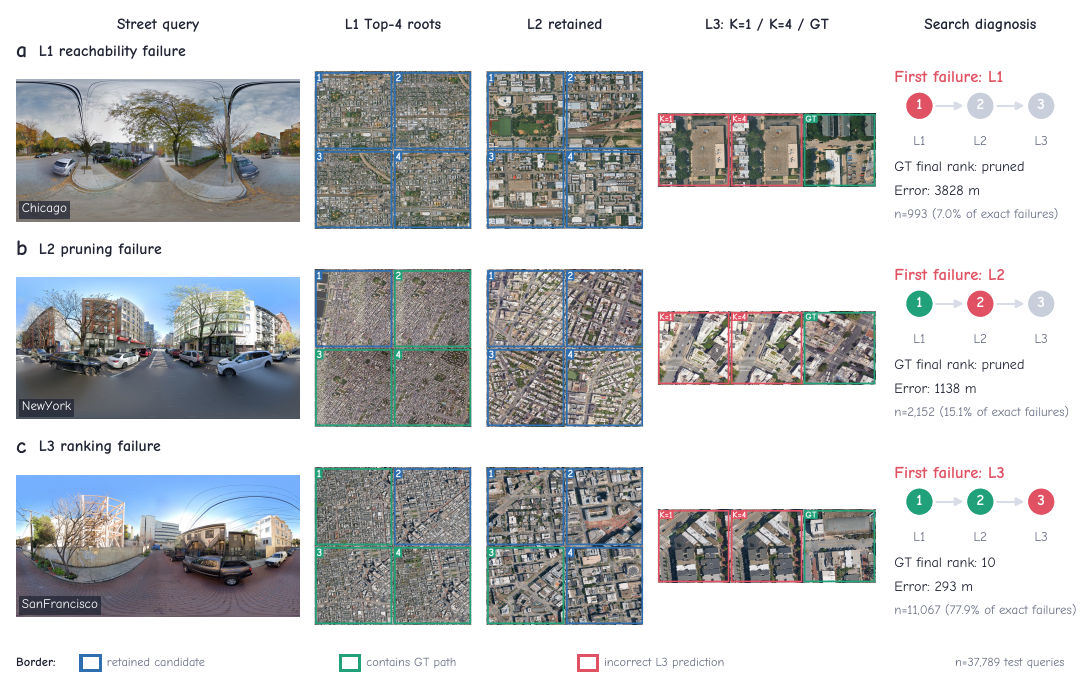}
  \caption{Failure modes of GeoMoE Beam-4 retrieval with PRC on VIGOR-M. Blue,
  green, and rose borders denote retained candidates, candidates on the
  ground-truth (GT) path, and incorrect L3 predictions, respectively.
  \textbf{a}, L1 reachability failure: retained roots cannot generate the GT L2
  tile. \textbf{b}, L2 pruning failure: the GT L2 tile is removed before L3
  expansion. \textbf{c}, L3 ranking failure: the GT L3 tile reaches the final
  candidate set but is not ranked first.}
  \label{supp:fig_failure_cases}
\end{figure*}

Figure~\ref{supp:fig_failure_cases} illustrates the three mechanisms. In panels
\textbf{a} and \textbf{b}, the ground-truth tile is absent from the final
candidate set, so PRC cannot recover the path. In panel \textbf{c}, candidate
generation succeeds but the ground-truth tile ranks tenth.

\section{Limitations}
\label{supp:limitations}

VIGOR-M covers four cities in the United States under a regular \(4\times4\)
hierarchy. The evaluation therefore does not establish transfer to broader
regions, irregular branching, nonuniform depth, or galleries with a different
distribution of observed cells. Validation across additional cities, seasons,
sensors, and land-use references remains necessary before treating the reported
efficiency--accuracy trade-off as geographically general. The intermediate-scale
benchmark is also method-level: the systems differ in backbone, model count, and
search protocol, and Just Zoom In receives a location oracle. Its greedy stage
hits are therefore not directly comparable to Beam-4 path retention.

GeoMoE is trained exclusively on observed cross-view image pairs and uses no
generative augmentation. Synthetic-data distillation and detector-guided
rewards have improved general-purpose image generation
\cite{Echo4o2025,RealGen2025}, while agentic systems incorporate multimodal
search, reasoning, or code-rendered intermediate canvases
\cite{MindBrush2026,GenClaw2026}. Whether such generators can produce
geographically faithful cross-view pairs, rather than merely plausible images,
and improve retrieval under region, season, or sensor shifts remains untested.

Both representation analyses are observational. The resolution diagnostic uses
only one checkpoint, and its two-dimensional PCA projection explains only a
small fraction of descriptor variance. Semantic categories are assigned post
hoc, while external land-use validation is restricted to high-confidence cells
in Washington, DC. The GeoMoE--DINOv2 probe contrast combines geolocalization
fine-tuning, multi-scale supervision, and the MoE architecture. The matched dense
MLJ retrieval control does not isolate this probe contrast; reduced scale
decodability cannot be assigned specifically to the MoE block without probing
that control.

Finally, the failure decomposition describes the locked Beam-4 protocol and
does not isolate beam-width or calibration effects. The representative panels
illustrate error categories rather than causal mechanisms.

\clearpage

\clearpage
\bibliography{aaai2027}

\end{document}